\pdfoutput=1

\documentclass[11pt]{article}

\usepackage{booktabs}
\usepackage[dvipsnames]{xcolor}
\usepackage[preprint]{acl}

\usepackage{times}
\usepackage{latexsym}
\usepackage{hyperref}
\usepackage{soul}
\usepackage[most]{tcolorbox}
\usepackage[T1]{fontenc}

\usepackage[utf8]{inputenc}

\usepackage{microtype}

\usepackage{inconsolata}

\usepackage{graphicx}
\usepackage{subcaption}
\usepackage{multirow}
\usepackage{amssymb}
\usepackage{cleveref}
\usepackage{enumitem}
\usepackage{subcaption}
\title{The Impact of Large Language Models in Academia: \\ from Writing to Speaking}

\author{
 \textbf{Mingmeng Geng\textsuperscript{1}\thanks{Correspondence to: Mingmeng Geng (\texttt{mgeng@sissa.it})}},
 \textbf{Caixi Chen\textsuperscript{2}},
 \textbf{Yanru Wu\textsuperscript{2}},
 \textbf{Dongping Chen\textsuperscript{2 3}},
\\
 \textbf{Yao Wan\textsuperscript{2}},
 \textbf{Pan Zhou\textsuperscript{2}}\\
 \textsuperscript{1}
International School for Advanced Studies (SISSA), \\
 \textsuperscript{2}
 Huazhong Univerisity of Science and Technology,
 \textsuperscript{3}
 University of Washington}

\begin{document}
\maketitle
\begin{abstract}
Large language models (LLMs) are increasingly impacting human society, particularly in textual information. Based on more than 30,000 papers and 1,000 presentations from machine learning conferences, we examined and compared the words used in writing and speaking, representing the first large-scale study of how LLMs influence the two main modes of verbal communication and expression within the same group of people. Our empirical results show that LLM-style words such as \textit{``significant''} have been used more frequently in abstracts and oral presentations. The impact on speaking is beginning to emerge and is likely to grow in the future, calling attention to the implicit influence and ripple effect of LLMs on human society.
\end{abstract}

\section{Introduction}

The development and popularity of large language models (LLMs) \citep{openai_gpt4o_2024, anthropic2024claude35, OpenAI2023O1Preview} have alerted more and more researchers to the impact of LLMs on human society. In this paper, we focus on the impact of LLMs in academia, especially on writing and speaking.

While the rapid increase in usage and impact of LLMs have been demonstrated in academic papers  \citep{liang2024mapping,geng2024chatgpt}, few studies have addressed the influence of LLMs beyond writing. Only recently, a preprint pointed out the impact of LLMs on the words used in speaking, as collected in YouTube videos \citep{yakura2024empirical}. The similarities and differences in how writing and speaking are influenced, particularly for the same population, have not been explored. 

People can use LLMs to write emails or accomplish tasks other than paper writing, which changes their English expression and is reflected in their academic output at a later point in time. Just like the use of Google Translate can affect the English expression of non-native English speakers \citep{resende2021can}, a similar influence might be at play with LLM users and eventually influence even the way people speak.

Besides, detecting a mixture of machine-generated and human-written text is another difficulty being actively researched \citep{lee2022coauthor,gao2024llm}. Researchers have paid more attention to whether a piece of text is generated by LLMs, while the implicit impact of LLMs is often underestimated. Here we mean people do not directly use LLMs to create content but have been influenced through exposure to such content.

In the face of these challenges and gaps, our contributions are three-fold:
\begin{enumerate}[itemsep=0pt, leftmargin=*]
    \item We are the first to analyze and compare the impact of LLM on the writing and speaking in the same group of people.
    \item We propose a simplified simulation-based method for estimating LLM impact.
    \item We are calling attention to the implicit impact of LLMs, as the words used in the machine learning conference presentations show signs of being influenced by LLMs.
\end{enumerate}

\section{Data and Methods}
\label{methods}

\begin{figure*}[t]
    \centering
    \includegraphics[width=\textwidth]{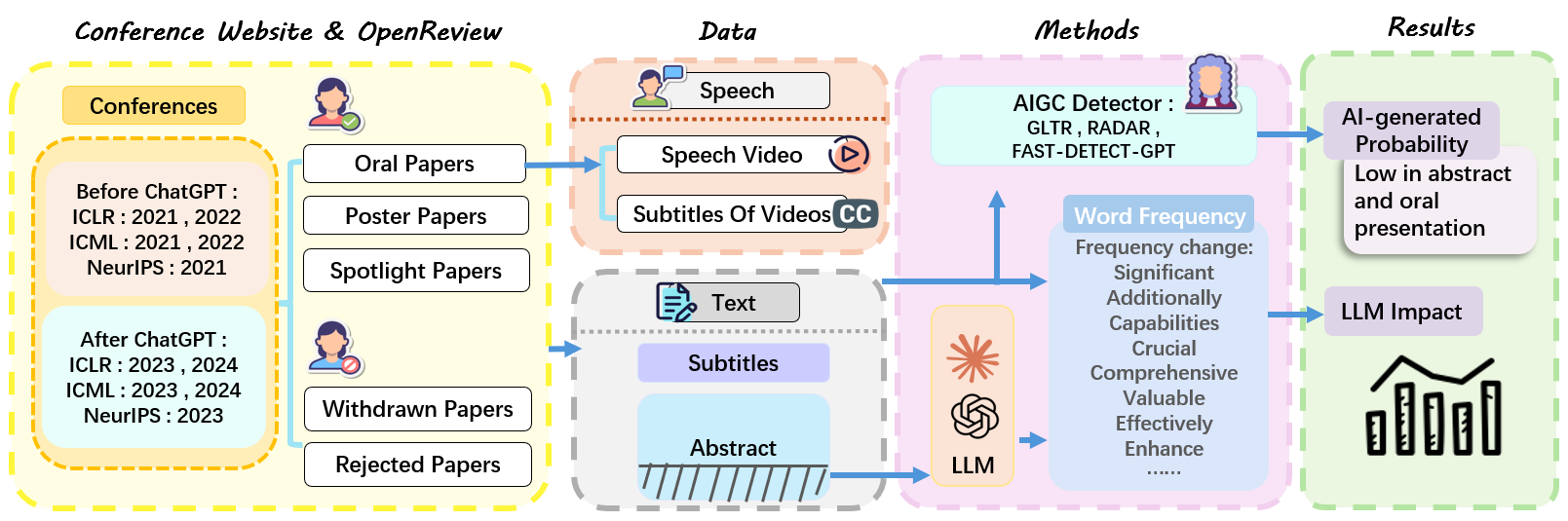}
    \vspace{-1.5em}
    \caption{Overview of processing and analysis.}
    \label{overview}
\end{figure*}

\subsection{Datasets}
 To better explore and compare how the ``same'' people are affected in writing and speaking by LLMs, we crawled presentations and meta-information of papers from three machine learning conferences. The abstracts of papers rather than the full papers were used in the analysis, as the former are more representative. More than 30,000 abstracts and 1,000 talks were collected, detailed in Appendix~\ref{data_experiments}. 
 
 Then we use some AIGC (AI-Generated Content) detectors or analyze the changes in word frequency. The whole process is shown in Figure~\ref{overview}. 

\subsection{Word Frequency Analysis}
\label{W_I}
Given that word frequencies are always changing, the issue of noise cannot be ignored. To reduce the error caused by the randomness of word usage, the target words are considered as a group, denoted as $W_I=\{w_i \mid i\in I \}$, where $i$ is the frequency ranking in the corresponding dataset.

For a group of words $W_{I_0}$, the control group with shift $n$ is defined as $W_{I_n}=\{w_{i+n} \mid i\in I_0 \}$. Given a corpus $S$, the corresponding frequency $F_n(S)$ is $F_n (S) = \sum_{w \in I_n} f_w (S)$, where $f_w$ means the frequency of word $w$ in set $S$. The frequency ratio between two different corpus is $R_n(S,S') = \frac{F_n (S)}{F_n (S')}$. 

Constructing control groups to analyze changes in word frequency has been used before \citep{matsui2024delving,yakura2024empirical}. In this paper, the words in each group have roughly the same frequency based on the ranking in the dataset, which shows whether the change in frequency of the target words is unusual.

\subsection{LLM Simulations and Impact Estimation}
Some researchers have estimated the impact of LLMs by excess vocabulary only \citep{kobak2024delving}, but the words in the abstract are also related to the topic of papers, and the hot topics of machine learning conferences change frequently. Therefore, it is also helpful to perform LLM simulations, and compare texts before and after processing for a reliable estimation of LLM impact.

If the frequency of word $i$ is $f(S_1)$ and $f(S_2)$ in a corpus before and after LLM processing, the frequency change rate ${r}_{i}$ is estimated as $\hat{r}_{i} = \frac{f(S_2)-f(S_1)}{f(S_1)}$. Then for the \textit{``proportion''} (impact) of LLMs texts $\eta(S)$, the following equation is a simplified and direct version of the method proposed by \citet{geng2024chatgpt},
\begin{equation}
        f^d_{i}(S) - f^*_{i}(S) =  \eta(S)f^*_{i}(S)\hat{r}_{i} + \delta_{i}(S) \label{f_ori}
\end{equation}
where $f^d_{i}(S)$ represents the frequency of word $i$ in the set of texts $S$, $f^*_{i}(S)$ represents the one if LLMs do not affect writing abstracts, and $\delta_{i}(S)$ is a noise term.

The estimate of LLM impact given by Ordinary Least Squares (OLS) is expressed as
\begin{equation}
    \hat{\eta}(S) = \frac{\sum_{i \in I} (f^d_{i}(S) - f^*_{i}(S))f^*_{i}(S)\hat{r}_{i}}{\sum_{i \in I} (f^*_{i}(S)\hat{r}_{i})^2} \label{eta_hat}
\end{equation}
where $I$ is the set of words used for estimation, and different $I$ give us different estimates.

\section{Results}
\subsection{AIGC Detectors}
Several AIGC detectors\footnote{We do not use \href{https://gptzero.me/}{GPTZero} because the size of its context window is smaller than the length of oral presentation.} like
Fast-DetectGPT \citep{bao2023fast}, GLTR \citep{gehrmann2019gltr} and RADAR \citep{Hu2023RADARRA} are used to detect the probability that the abstracts and speeches are generated by AI. The results show no significant difference between the pre-ChatGPT and post-ChatGPT era, more details and results are in Appendix \ref{detector}.

Therefore, we dive deeper into fine-grained word frequency analysis, as researchers have discovered that the frequency of some words increased rapidly across academic papers in different disciplines after the end of 2022 \citep{liang2024monitoring}.

\subsection{Changes in Word Frequencies}
We found a similar trend for some words in the abstracts of conference papers. For example, the 8 words (\textbf{significant}, \textbf{crucial}, \textbf{effectively}, \textbf{additionally}, \textbf{comprehensive}, \textbf{enhance}, \textbf{capabilities}, \textbf{valuable}), listed as examples by \citet{geng2024chatgpt}, are much more often observed in the abstracts of ICLR 2024 accepted papers than before, as shown in Figure \ref{wc_iclr_poster} (and Figure \ref{wc_iclr_supp} in the appendix). Besides, Figure \ref{wc_iclr} indicates that the frequency sums of the 8 words are \textit{``significantly''} higher in the ICLR 2024 abstracts than in abstracts of 2021 and 2022, in all three categories of accepted papers, as well as in the \textit{reject} paper and \textit{withdrawn} papers. The results also suggest the correlation between the use of these words and their destination in ICLR, with a lower frequency in \textit{oral} papers and a higher frequency in \textit{withdrawn} papers.

\begin{figure}[t]
    \centering
    \begin{subfigure}[b]{0.22\textwidth}
        \includegraphics[width=\textwidth]{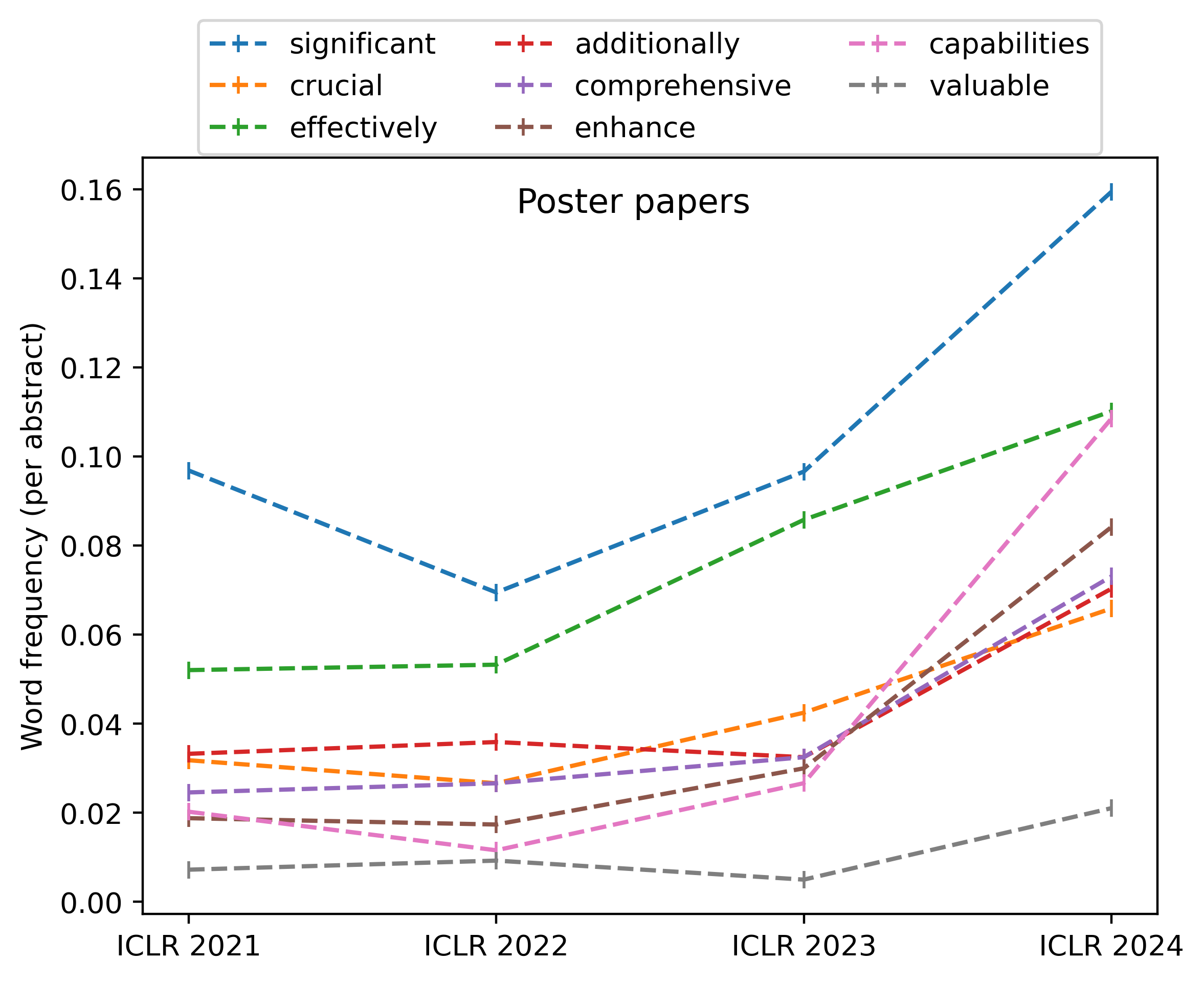}
        \caption{In abstracts of ICLR \textit{poster} papers.}
        \label{wc_iclr_poster}
    \end{subfigure}
    \hfill
    \begin{subfigure}[b]{0.22\textwidth}
        \includegraphics[width=\textwidth]{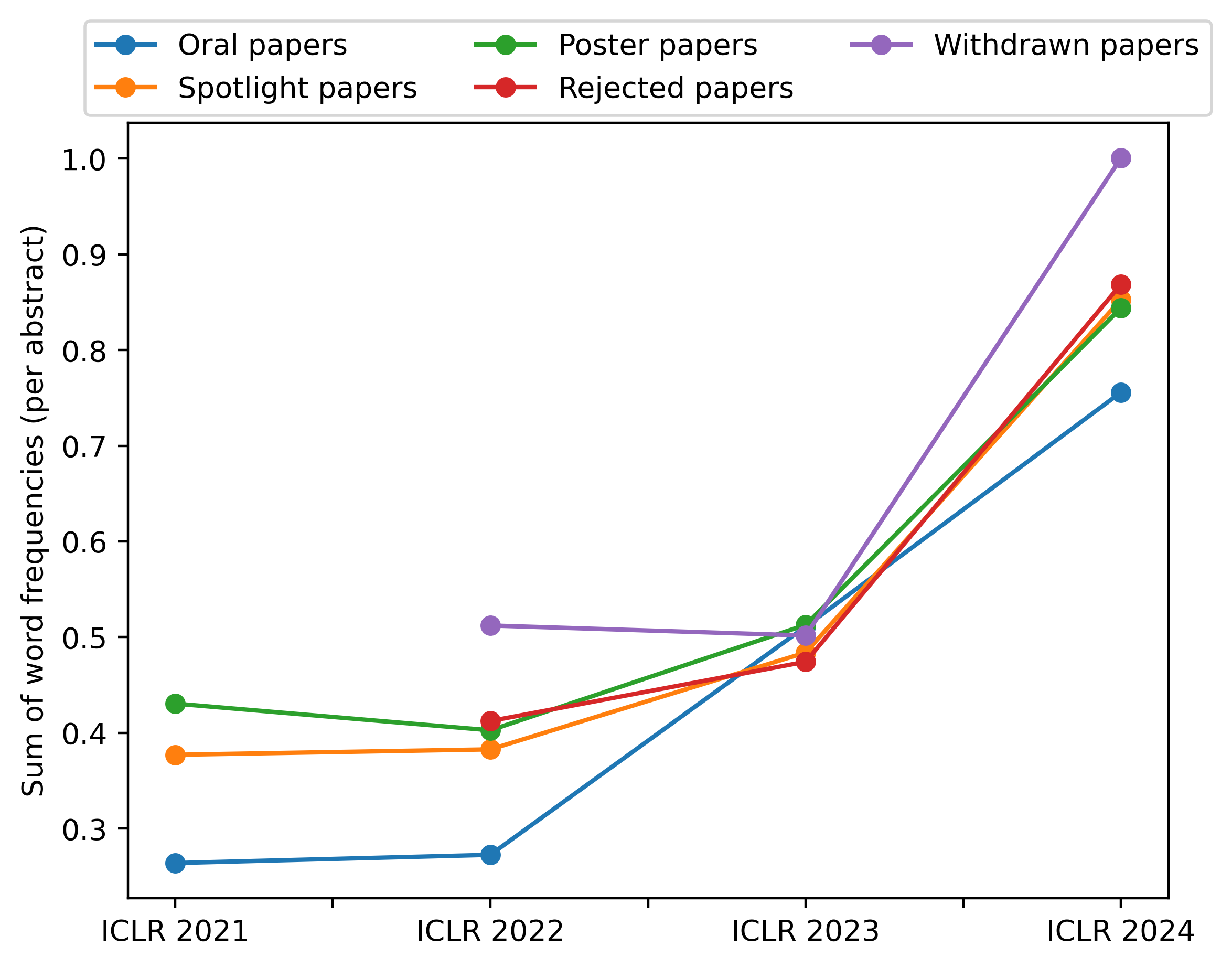}
        \caption{In abstracts of ICLR papers (sum).}
        \label{wc_iclr}
    \end{subfigure}
    \begin{subfigure}[b]{0.22\textwidth}
        \includegraphics[width=\textwidth]{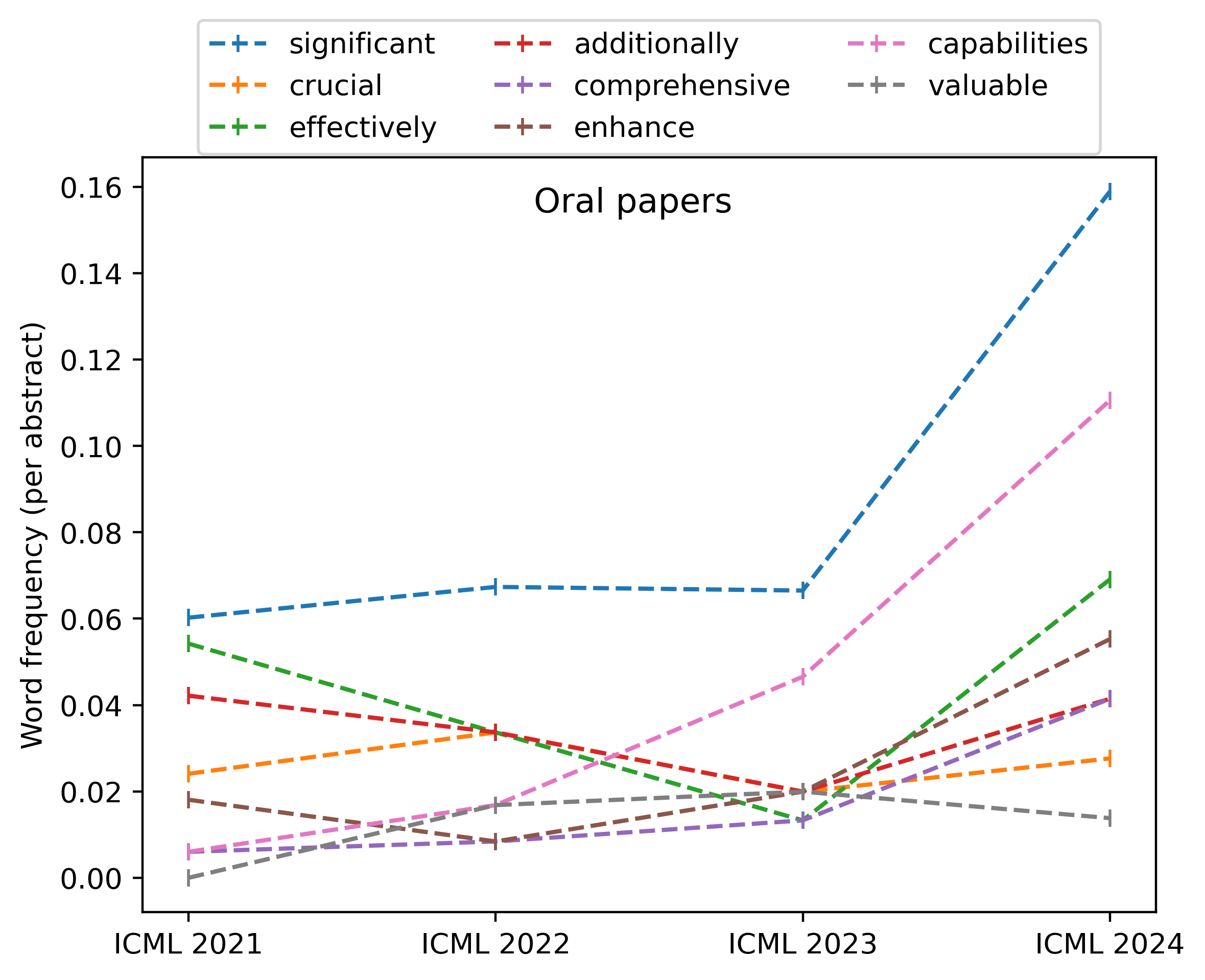}
        \caption{In abstracts of ICML \textit{oral} papers.}
        \label{wc_icml_oral}
    \end{subfigure}
    \hfill
    \begin{subfigure}[b]{0.22\textwidth}
        \includegraphics[width=\textwidth]{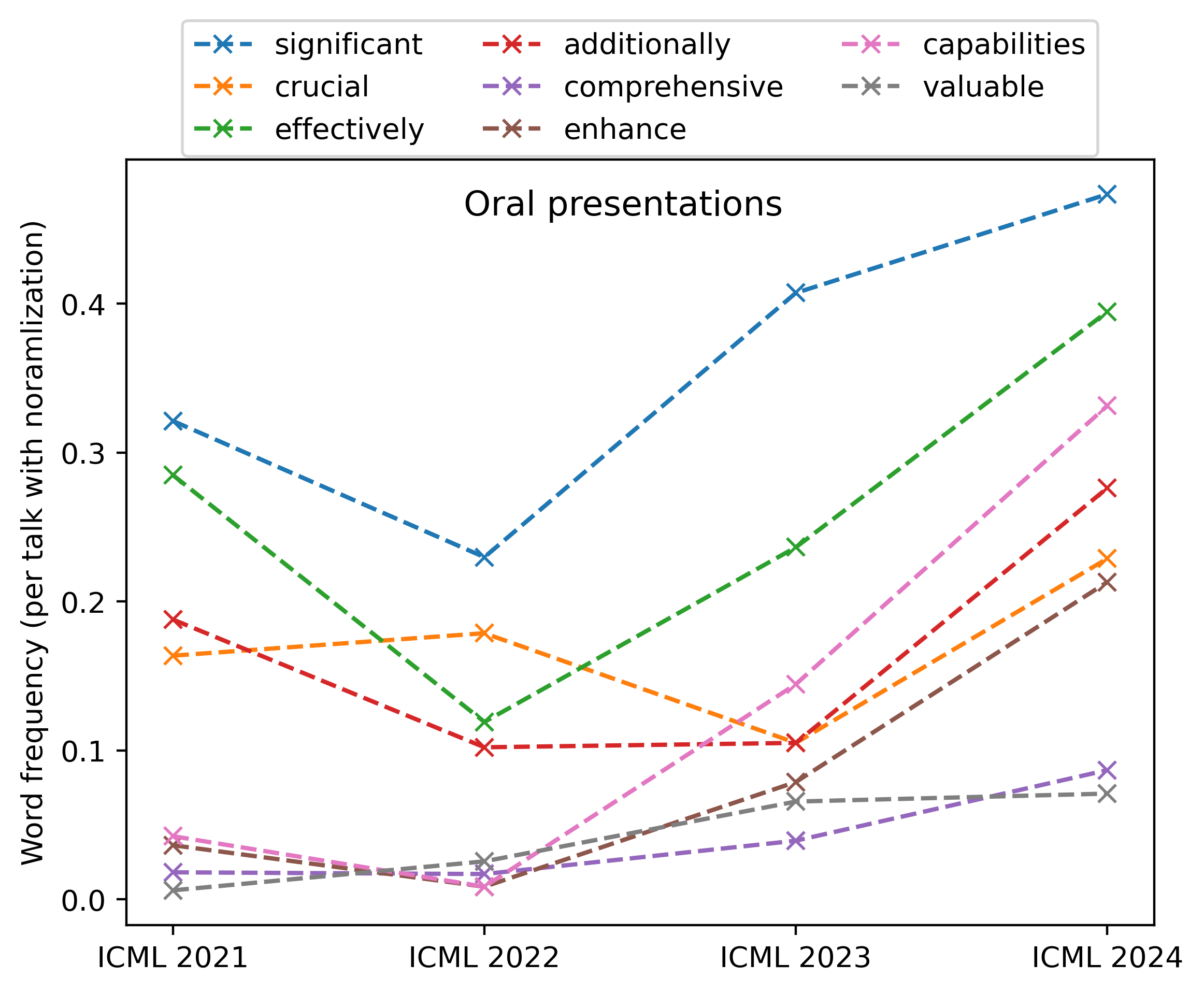}
        \caption{In presentations of ICML \textit{oral} papers.}
        \label{wc_icml_speech}
    \end{subfigure}    
    \caption{Word frequency in abstracts and presentations.}
    \label{wc_iclr_icml}
    \vspace{-1em}
\end{figure}

Although these conference papers are mostly submitted to the arXiv as well, their average quality is higher than the same type of arXiv papers. These eight words shown above were taken together for better comparison and were not selected based on the data in this paper, reconfirming the LLM impact in academic research. 

We further compared word frequency used in the abstracts and speeches of the \textit{oral} papers. Since the time for presentations varied from year to year, the number of words was normalized based on the total number of words used in 2021. The results for ICML \textit{oral} papers and presentations are presented in Figure \ref{wc_icml_oral} and Figure \ref{wc_icml_speech}, which shows that these example words are more frequently used in the abstracts of papers after 2022. Words in the speeches share the same trend as well, though not as strongly as in the abstracts. 

\subsection{Distribution of Frequency Ratios}
\begin{figure}[t]
    \centering
    \begin{subfigure}[b]{0.22\textwidth}
        \includegraphics[width=\textwidth]{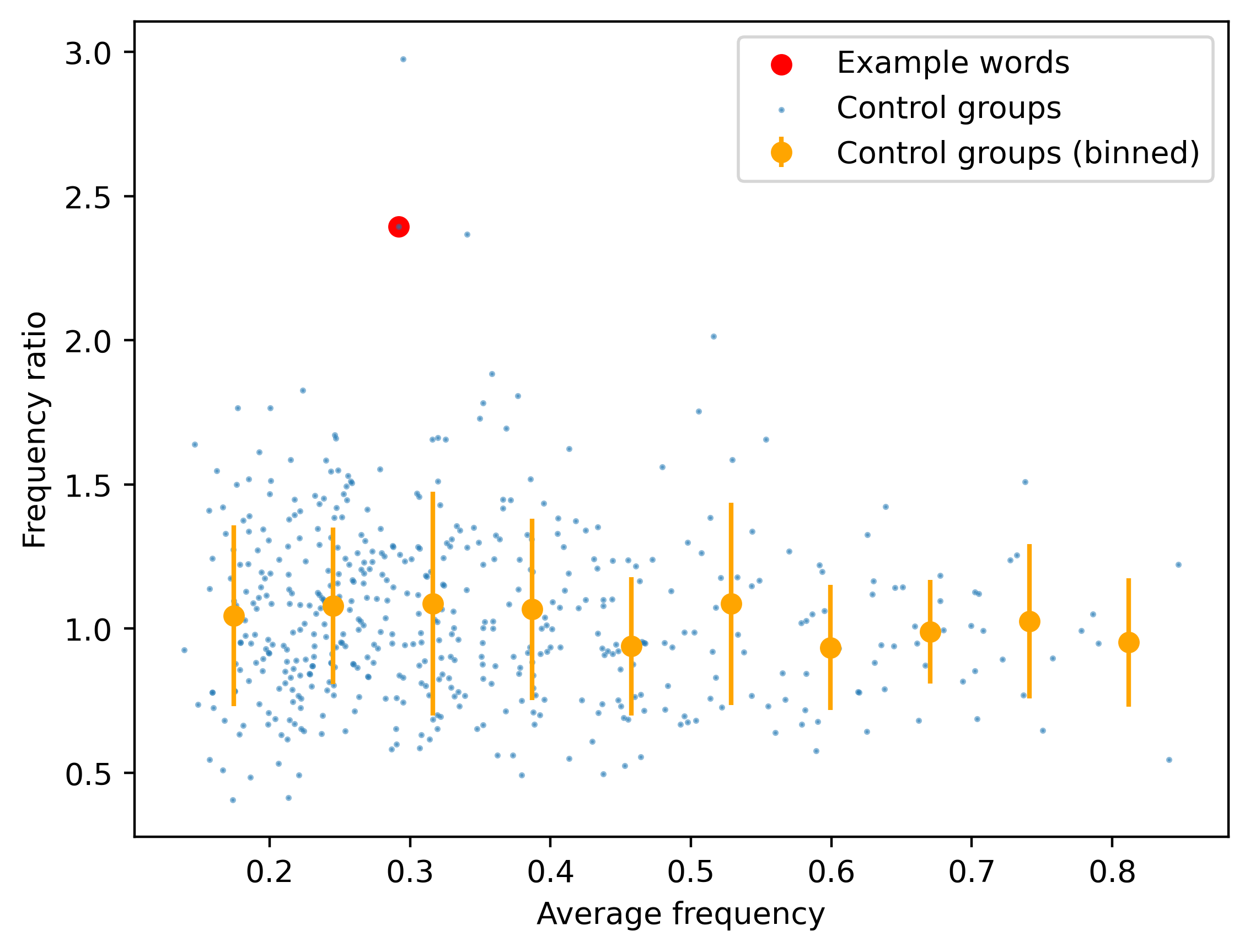}
        \caption{Frequency ratio in abstracts of \textit{oral} papers.}
    \end{subfigure}
    \hfill
    \begin{subfigure}[b]{0.22\textwidth}
        \includegraphics[width=\textwidth]{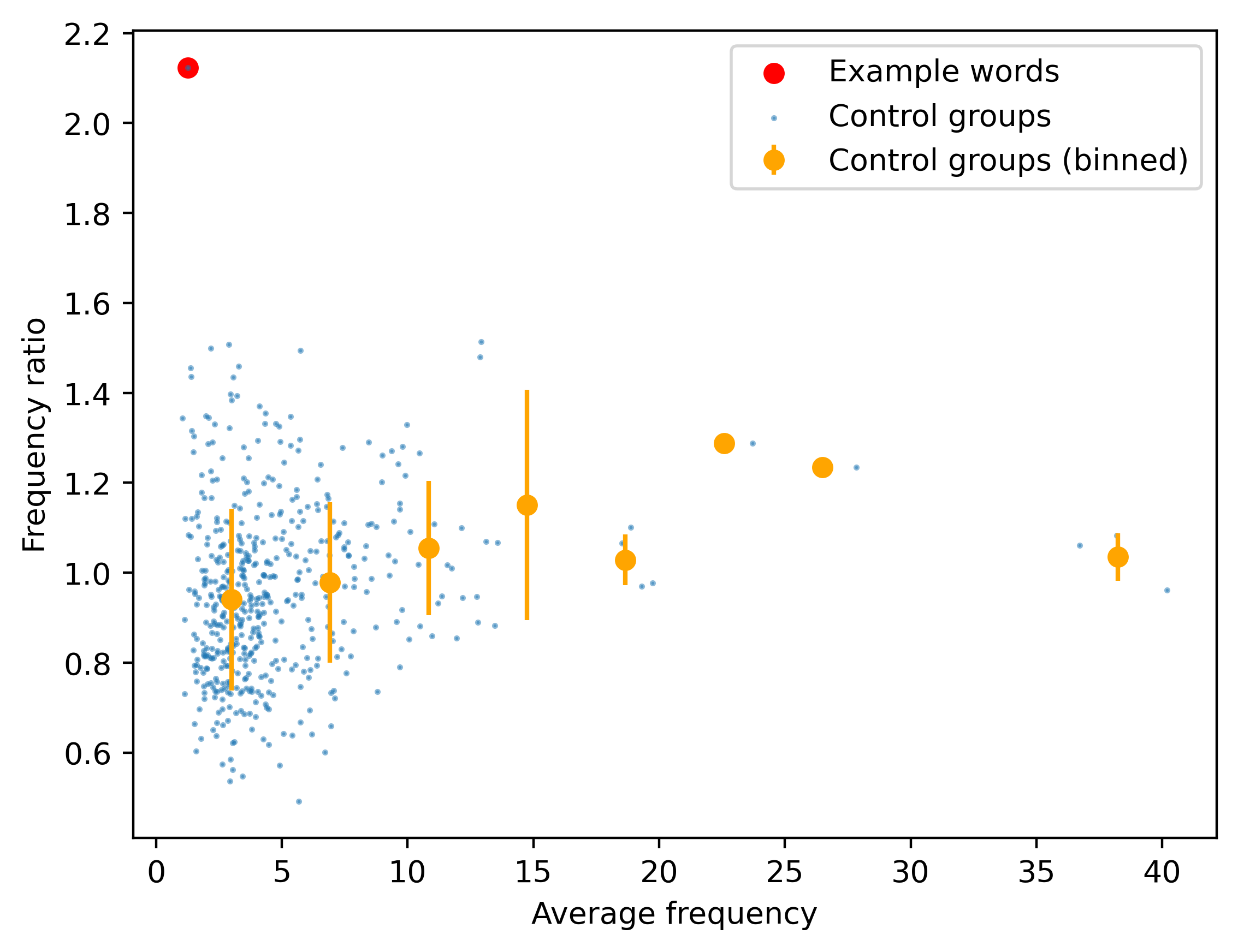}
        \caption{Frequency ratio in presentations of \textit{oral} papers.}
    \end{subfigure}
    \caption{Word frequency ratio in abstracts and talks of ICML \textit{oral} papers. The error bars represent one standard deviation in each bin.}
    \label{ratio_icml}
\end{figure}

The word frequencies in the abstracts of poster papers in NeurIPS from 2021 to 2023 were used for ranking $i$ the words to form the control group $W_I$ with the shift as defined in section \ref{W_I}. The frequency ratio $R_n(S, S')$ for the abstracts and talks of the ICML oral papers in 2024 compared to those from 2021 to 2022 are shown in Figure \ref{ratio_icml}, with shift $n$ from $-250$ to $250$ forming the control groups. 

The frequency ratios of the example word group are 3.4 standard deviations and 5.8 standard deviations away from the mean within the bin in abstracts and speeches, respectively. The vast majority of word groups do not have as much frequency change as the example word group.

\subsection{LLM Simulations}
\begin{figure}[t]
    \centering
    \begin{subfigure}[b]{0.22\textwidth}
        \includegraphics[width=\textwidth]{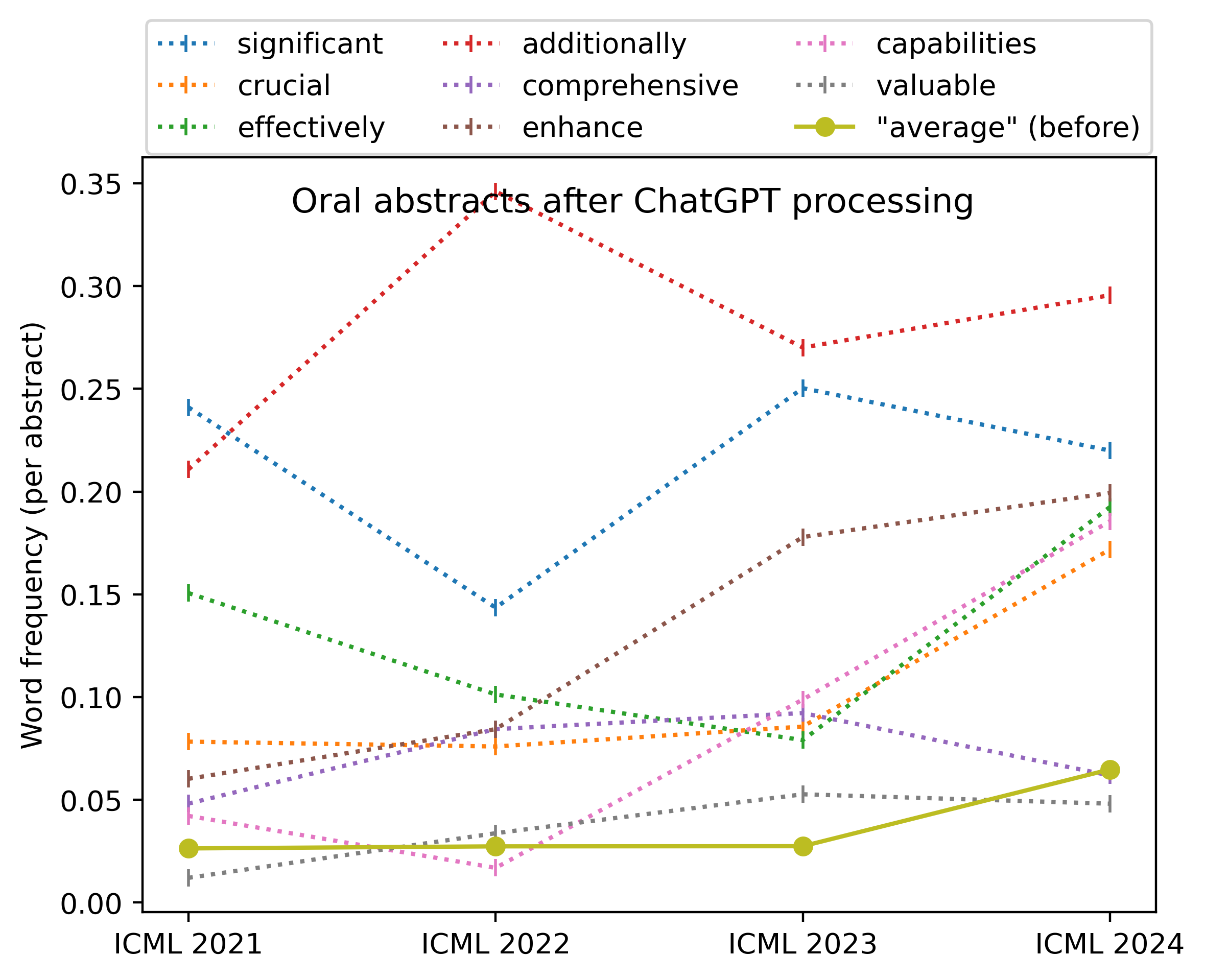}
        \caption{Word frequency after ChatGPT processing.}
        \label{wc_icml_oral_sim}
    \end{subfigure}
    \hfill
    \begin{subfigure}[b]{0.22\textwidth}
        \includegraphics[width=\textwidth]{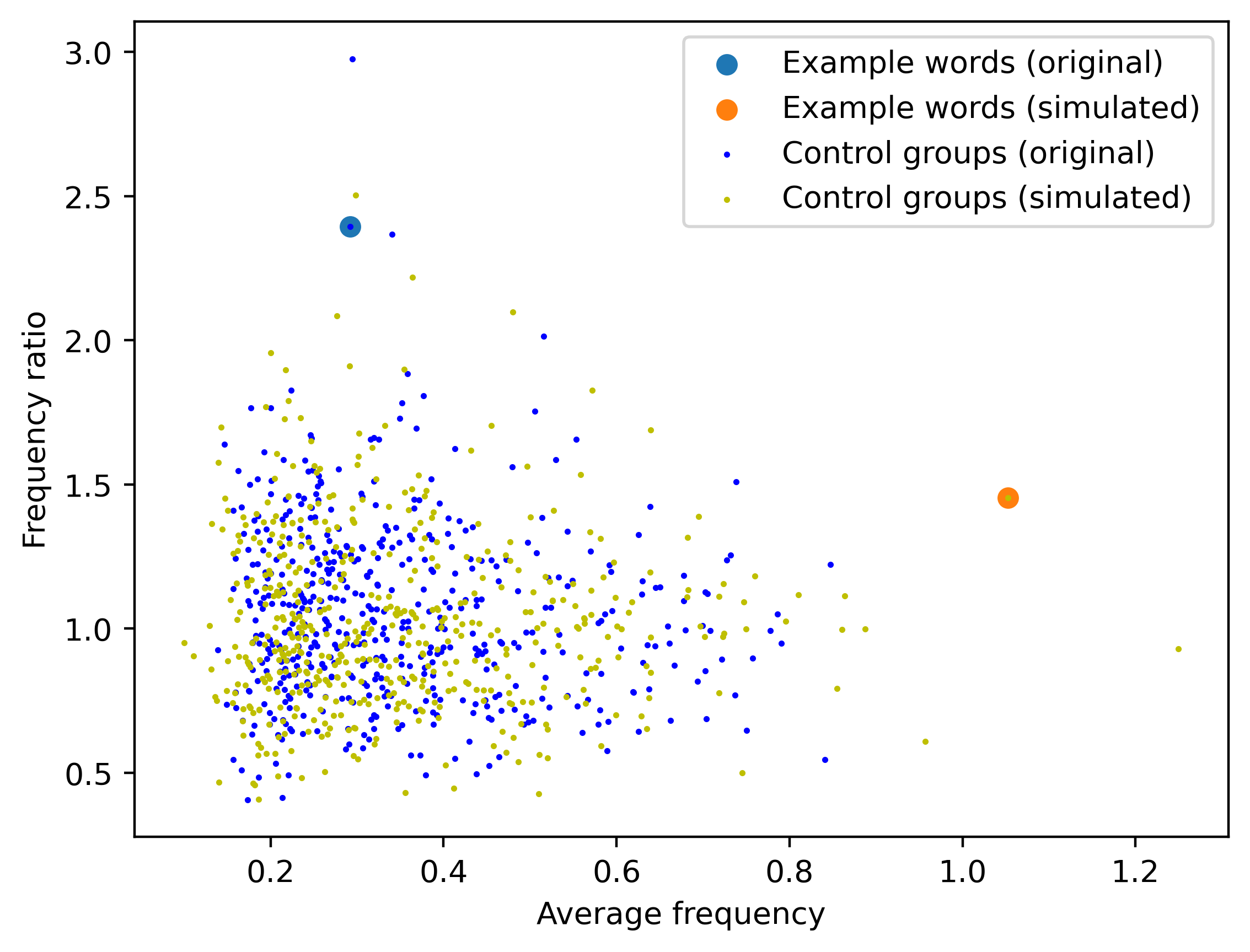}
        \caption{Word frequency ratio between simulated abstracts and original abstracts.}
        \label{ratio_icml_oral_compare}
    \end{subfigure}
    \caption{Word frequency in abstracts of ICML \textit{oral} papers.}
    \label{sim_compare}
\end{figure}

The simulations of LLM were performed on GPT-3.5 with a simple prompt: \textit{``Revise the following sentences''}.

The word frequency analysis on the abstracts revised by ChatGPT is in Figure \ref{wc_icml_oral_sim}, obtained after the same calculations as for Figures \ref{wc_iclr_poster} and \ref{wc_icml_oral}. The frequency of these words has all increased after ChatGPT processing, which reconfirms that ChatGPT favors these words. The comparison of simulated and original data in Figure \ref{ratio_icml_oral_compare} tells us that words sensitive to ChatGPT processing are in the minority.

Based on our results and previous papers, it is hard to believe that the change in word frequency revealed in Figure \ref{wc_iclr_icml} is a coincidence. Further, the results above have illustrated the correlation between word frequency changes and word preferences of LLMs.

\begin{figure}[t]
    \centering
    \begin{subfigure}[b]{0.45\textwidth}
        \includegraphics[width=\textwidth]{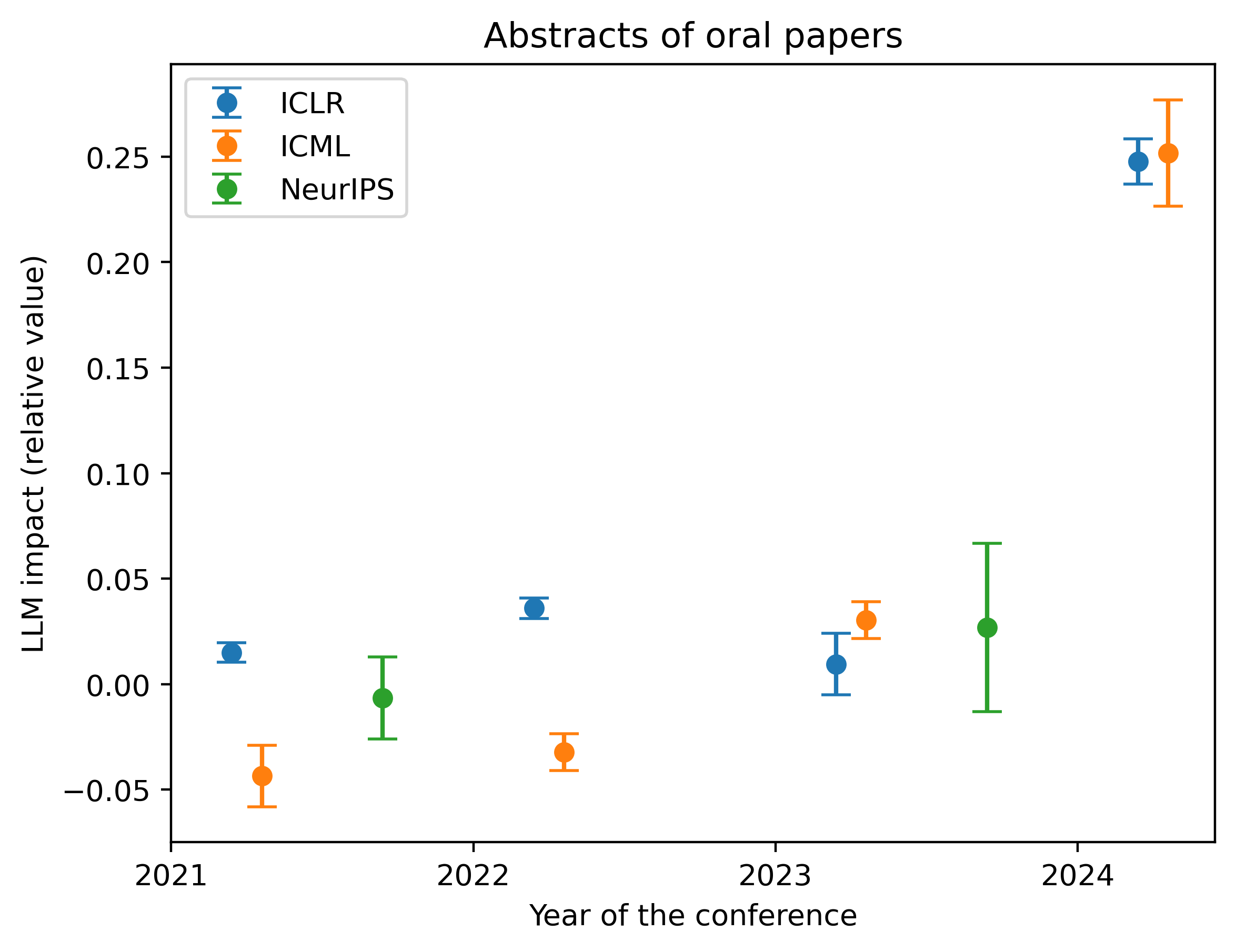}
    \end{subfigure}
    \hfill
    \begin{subfigure}[b]{0.45\textwidth}
        \includegraphics[width=\textwidth]{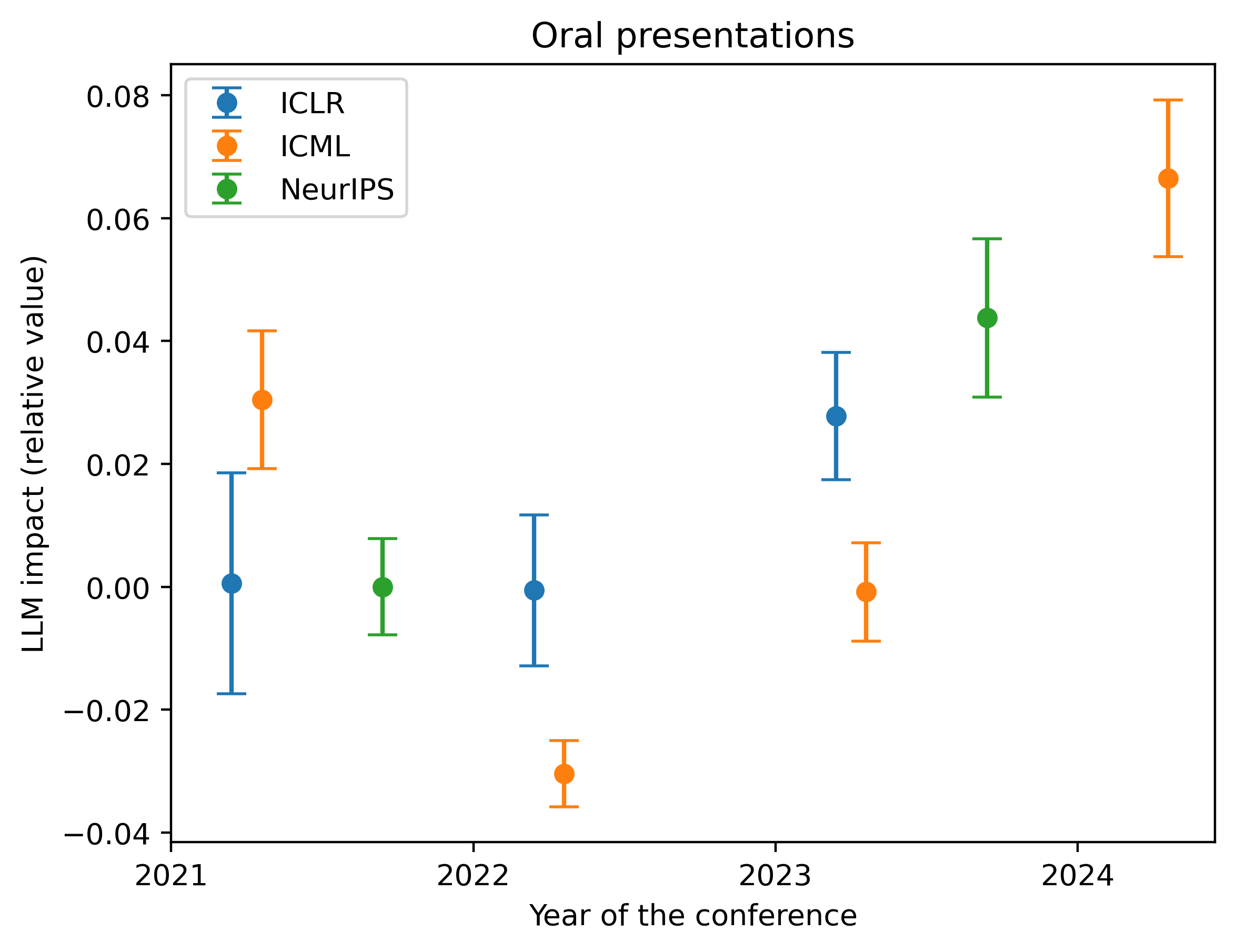}
    \end{subfigure}
    \caption{Estimation of LLM's impact on oral papers. Error bars indicate the standard deviation of the estimates using different groups of words.}
    \label{llm_impact}
    \vspace{-1em}
\end{figure}

\subsection{LLM Impact Estimations}
The ChatGPT-modified abstracts of spotlight (poster) papers from ICML 2022 were utilized to calculate $\hat{r}_i$ and LLM impact estimations. And $f^*_{i}(S)$ is approximated by the word frequency of the abstracts of poster or spotlight papers in 2021 for each conference, after normalization to have the same total number of words. To reduce the effect of topic-specific words, only the words that ranked in the top 10,000 of the Google Ngram dataset\footnote{\url{https://www.kaggle.com/datasets/wheelercode/english-word-frequency-list}} were considered in the calculations. Finally, hundreds of words are used for estimating, not just the eight previously exemplified.

Because of the lack of presentation data before 2021, the $f_i^*(S)$ of speeches were also approximated with the contents of the abstracts. Estimates of LLM impact based on different different word selection criteria determined by $f_i^*(S)$ and $\hat{r}_i$ are shown in in Figure \ref{llm_impact}. Since the words written and spoken are different, the means of the estimates for 2021 and 2022 were also calibrated to ``$0$'', with corresponding adjustments for 2023 and 2024. Not surprisingly, LLM impact increased in the abstracts for the 2024 conference. In the presentations, the estimate of LLM impact has also increased, but not as ``significantly" as in the abstracts.

Note that the LLM impact term here is a relative value, which is an estimate based on the output of a specific prompt via ChatGPT. Different prompts and different LLMs will have different outputs, and the implicit impact also plays a role.

\section{Conclusions and Discussions}
Our paper is the first to estimate the effect of LLMs on the same group of people in writing and speaking and one of the first to analyze the impact on human communication and beyond text. Although traces of LLMs are only nascent in the presentations of machine learning conferences, their usage and impact in writing have been illustrated by much literature and the results above and are becoming more and more profound in academia.

Our findings demonstrate the impact of LLMs in the abstracts of papers at top machine learning conferences. We also found that the words used in the talks were also influenced by LLMs, although this effect was less than in the abstracts.

Speech is one of the scenarios in which LLMs have an implicit impact (broadly), for it is usually safe to assume that the speaker is not using an LLM while presenting. In addition, because of the difference between written and spoken language, it is actually difficult to directly compare LLM impact on writing and speaking. 

The rapid increase in AI-generated content requires us to pay more attention, as synthetic data can lead to model collapse \citep{shumailov2024ai,guo2023curious,briesch2023large} and even knowledge collapse \citep{peterson2024ai}. In the near future, a paper considered likely to be the product of LLMs may be so only because the authors have read too many papers containing LLM-style sentences. The implicit impact of LLMs seems unstoppable.
\section*{Limitations}

LLMs have started a paradigm revolution in AI and transformed the game completely. While the discussion of the social impact of LLMs began long before the storm hit \citep{solaiman2019release}, it took some time to really \textit{``delve into''} it.

There are many sources of implicit impacts, and it is also difficult to have a standard definition. For instance, the titles of papers \citep{matsui2024delving,kobak2024delving,astarita2024delving} analyzing the use of LLMs in academic papers begin with \textit{``delving into''} (one of ChatGPT's signature words). This fits our definition of implicit impact in the broad sense, but some argue that it's just a means for authors to get readers' attention. 

It is also true that researchers may have prepared presentations or slides using LLMs, in which case their choice of wording may have been influenced. The sample sizes in speaking are not as large as in writing, but they are homogenized and representative.

We concentrated on the frequency of words and did not address other forms. The use of words reflects the most basic information, and some changes should have occurred in the way they are expressed.

\section*{Ethics Statement}
Our paper primarily focuses on LLM's influence on writing and speaking, and we must first declare that our research adheres to all applicable ethical standards. This study is intended to promote academic discussion and technological progress and all experiments are conducted in a strictly controlled environment. Our research encourages relevant developers to enhance the supervision of LLMs, thereby making them more trustworthy. We ensured that all datasets and benchmarks used in the study comply with their intended purposes and standards.

\bibliography{custom}

\clearpage
\appendix

\section{Related Works}
\label{related works}
\subsection{Impacts of LLMs}
The storm of LLMs also crossed the boundaries of language, as researchers began to explore the wide variety of competencies and applications of LLMs in various disciplines: math \citep{ahn2024large}, physics \citep{sun2024knowledge}, chemistry \citep{guo2023can}, social science \citep{geng2024large}, psychology \citep{demszky2023using}, cognitive science \citep{chen2024self}, and almost all corners of science. In the process of interacting with LLMs, people's opinions may also change \citep{costello2024durably}.

We know that many scientists are using LLMs, but there are not many papers with quantitative estimates of the impact of LLMs in the scientific community. Along with academic writing \citep{liang2024mapping,geng2024chatgpt}, LLMs have been found to be used in peer review \citep{liang2024monitoring,latona2024ai}, and it's hard to stop researchers from using LLMs \citep{bao2024there}.

\subsection{Machine Generated Text Detection}
Detection of machine-generated text has also begun much earlier with different approaches \citep{gehrmann2019gltr,bakhtin2019real,uchendu2020authorship}. More have approaches have been proposed later, including metric-based methods \citep{mitchell2023detectgpt,yang2023dna,bao2023fast}, model-based methods \citep{guo2023close,verma2023ghostbuster}, and benchmarks \citep{he2023mgtbench}. In addition, some papers also illustrate the ways to avoid LLM detection \citep{krishna2023paraphrasing,koike2023outfox}, which makes this game harder.

Detecting a mixture of machine-generated text and human-written text is another challenge. Recent research on \textit{``LLM-as-a-Coauthor''} \citep{zhang2024llm} explores the complex interactions between humans and LLMs in collaborative writing, investigating real-world applications of human-AI mixed text, moving beyond simple binary classification. In the face of mixed text \citep{zeng2024towards,ji2024detecting}, more methods have also been proposed in the literature for detection and estimation, e.g., through words \citep{liang2024monitoring} and styles \citep{gao2024llm}.

But with implicit impact and ripple effects from LLMs, the line between human-written and machine-generated text is indistinct, and monitoring LLM-style text may be more meaningful than detecting who the author is.

\newpage
\section{Data Collection and Experiments Setups}
\label{data_experiments}

\subsection{Data Collection}
The abstracts for ICML 2021 and 2022 were collected from the official website, and all other data were scraped from the \textit{OpenReview} platform. Since ICLR 2023 didn't distinguish \textit{oral} and \textit{spotlight}, we consider the \textit{notable top 5\%} papers and \textit{notable top 25\%} papers as them, respectively. As for NeurIPS 2022, accepted papers were not further classified. In addition, there are more than 10,000 rejected and withdrawn papers in ICLR from 2022 to 2024.

The videos of the oral presentation were collected from the official websites of ICLR, ICML, and NeurIPS. As NeurIPS 2022 did not feature oral papers, the videos of this year's NeurIPS were skipped. The videos of the ICLR 2024 oral presentation haven't been available online. To extract speech from the videos, we used the subtitle feature provided by each website. Using the Internet Download Manager (IDM) extension in Google Chrome, we downloaded all subtitle files in vtt format, which were converted into text via Python later.
\begin{table*}[h]
\centering
\caption{Statistics and source data link of each conference analyzed in our paper. Missing data are marked with \textit{``*''}.}
\label{tab:conference-stats}
\scalebox{0.8}{
\begin{tabular}{lccccc}
\toprule[1.5pt]
Conf. & Year & Oral & Spotlight & Poster & Source Link \\
\midrule
ICML & 2021 & 166 & 1017 & * & \url{https://icml.cc/Conferences/2021/Schedule} \\
& 2022 & 118 & 1115 & * & \url{https://icml.cc/Conferences/2022/Schedule} \\
& 2023 & 155 & * & 1673 & \url{https://openreview.net/group?id=ICML.cc/2023/Conference} \\
& 2024 & 144 & 191 & 2275 & \url{https://openreview.net/group?id=ICML.cc/2024/Conference} \\ \midrule
NeurIPS & 2021 & 60 & 284 & 2286 & \url{https://openreview.net/group?id=NeurIPS.cc/2021/Conference} \\
& 2022 & * & * & \textit{\underline{2671}} & \url{https://openreview.net/group?id=NeurIPS.cc/2022/Conference} \\
& 2023 & 67 & 378 & 2773 & \url{https://openreview.net/group?id=NeurIPS.cc/2023/Conference} \\  \midrule
ICLR & 2021 & 53 & 114 & 692 & \url{https://openreview.net/group?id=ICLR.cc/2021/Conference} \\
& 2022 & 55 & 175 & 864 & \url{https://openreview.net/group?id=ICLR.cc/2022/Conference} \\
& 2023 & \textit{{\underline{90}}} & \textit{\underline{281}} & 1201 & \url{https://openreview.net/group?id=ICLR.cc/2023/Conference} \\
& 2024 & 86 & 367 & 1807 & \url{https://openreview.net/group?id=ICLR.cc/2024/Conference} \\
\bottomrule[1.5pt]
\end{tabular}}
\end{table*}

\begin{figure*}[ht]
    \centering
    \begin{subfigure}{0.45\textwidth}
        \includegraphics[width=\textwidth]{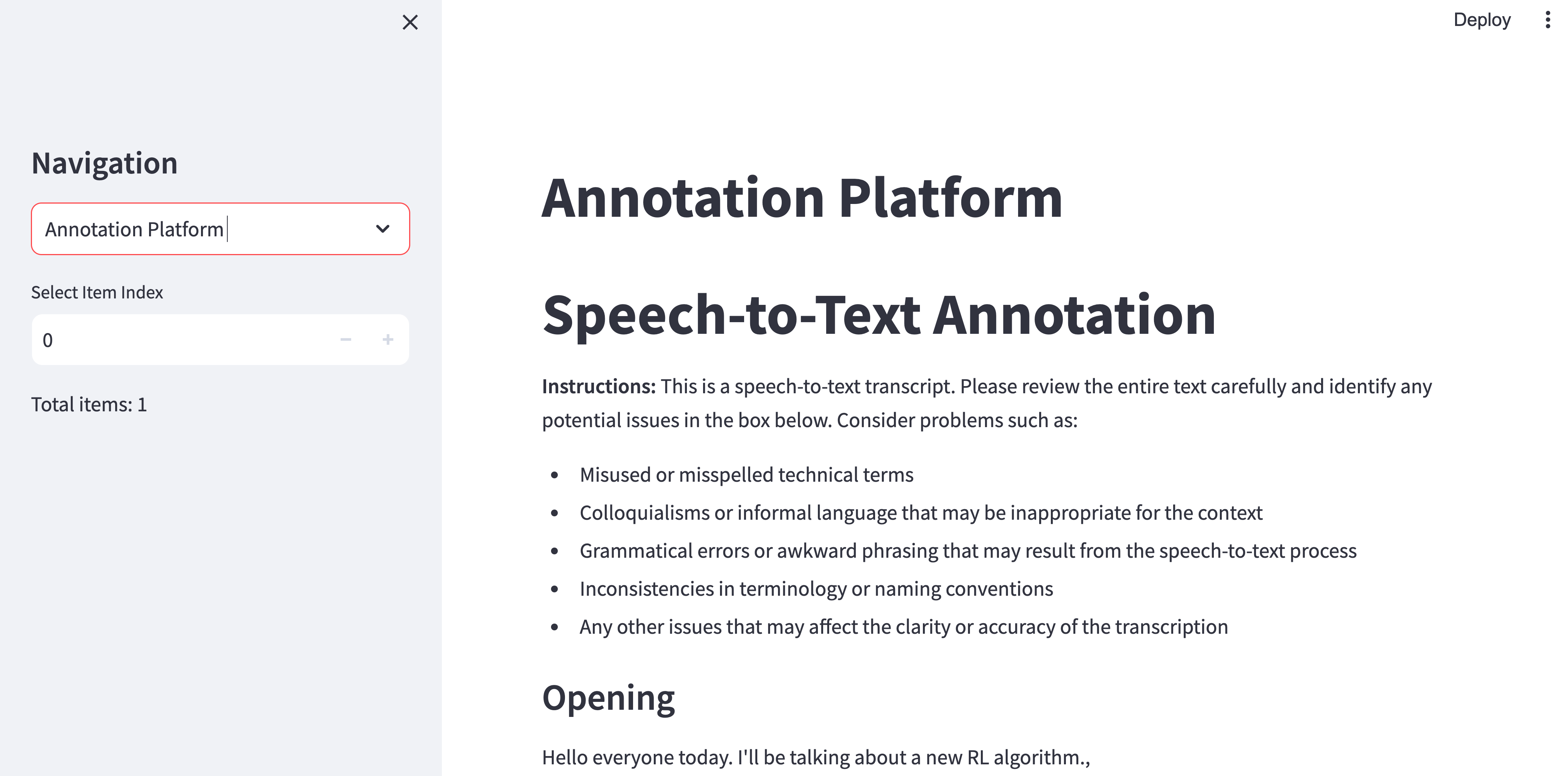}
        \caption{Instruction for annotators.}
    \end{subfigure}
    \hfill
     \begin{subfigure}{0.45\textwidth}
        \includegraphics[width=\textwidth]{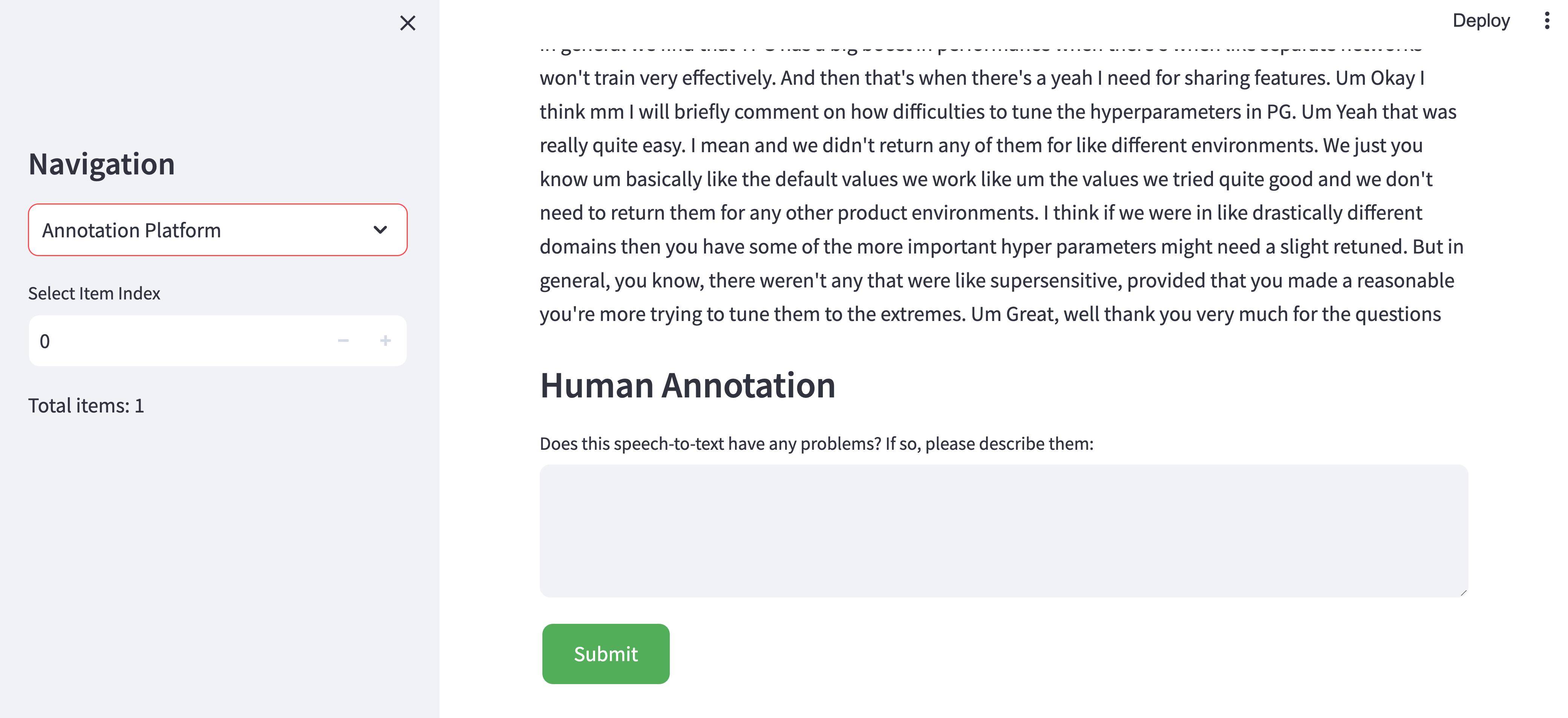}
        \caption{An example for human annotation.}
    \end{subfigure}
    \caption{Annotation details.}
    \label{fig:annotation}
\end{figure*}

\subsection{Human Annotation Details}
In this section, we provide details for human annotation.  Dataset of oral presentation and abstract are annotated with four author of this paper, with two are female and two are male. As acknowledged, the diversity of annotators plays a crucial role in reducing bias and enhancing the reliability of the benchmark. These annotators have knowledge in this domain, with different genders, ages, and educational backgrounds. To ensure the annotators can proficiently mark the data, we provide them with detailed tutorials, teaching them how to evaluate model responses more objectively. Specifically, they are required to give judgments without considering answer lengths, and certain names or positions of the response. Besides, we implement cross-validation between different annotators and conduct continuous monitoring to ensure they are
maintaining objectivity and fairness. We provide screenshots of the instruction and annotation in Figure \ref{fig:annotation}.

\subsection{LLM simulations}
\begin{itemize}
    \item model: gpt-3.5-turbo-0125
    \item temperature: 1
    \item seed: index of the abstract in the dataset 
    \item top\_p: 0.9
\end{itemize}
\subsection{Estimations of LLM impact}
The choice of words is based on the value of $f^*$ and $\hat{r}$.
\begin{itemize}
    \item $\frac{1}{f^*}$: 30, 40, 50, 60, 70, 80, 100, 150, 200, 500
    \item $\hat{r}$: 0.4, 0.5, 0.6, 0.7 (corresponding value of $\frac{\hat{r}+1}{\hat{r}^2}$)
\end{itemize}

\section{Supplementary Results}

\subsection{AIGC detector results}
\label{detector}
\begin{figure*}[t]
    \centering
    \includegraphics[width=\textwidth]{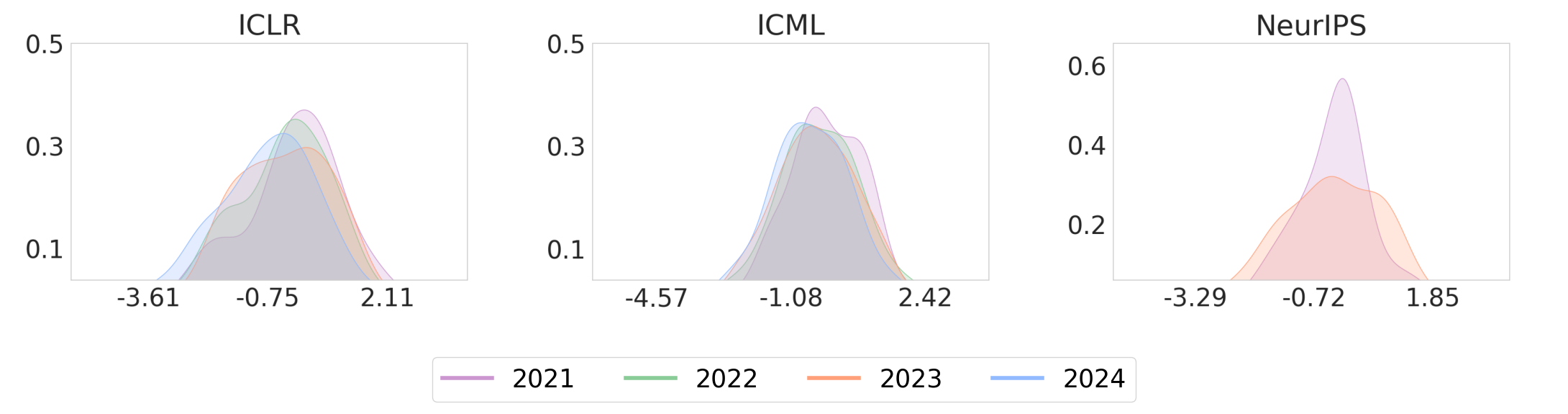}
    \caption{Criterion distribution in abstracts of oral works.}
    \label{Oral_abstract}
\end{figure*}

\begin{figure*}[t]
    \centering
    \includegraphics[width=\textwidth]{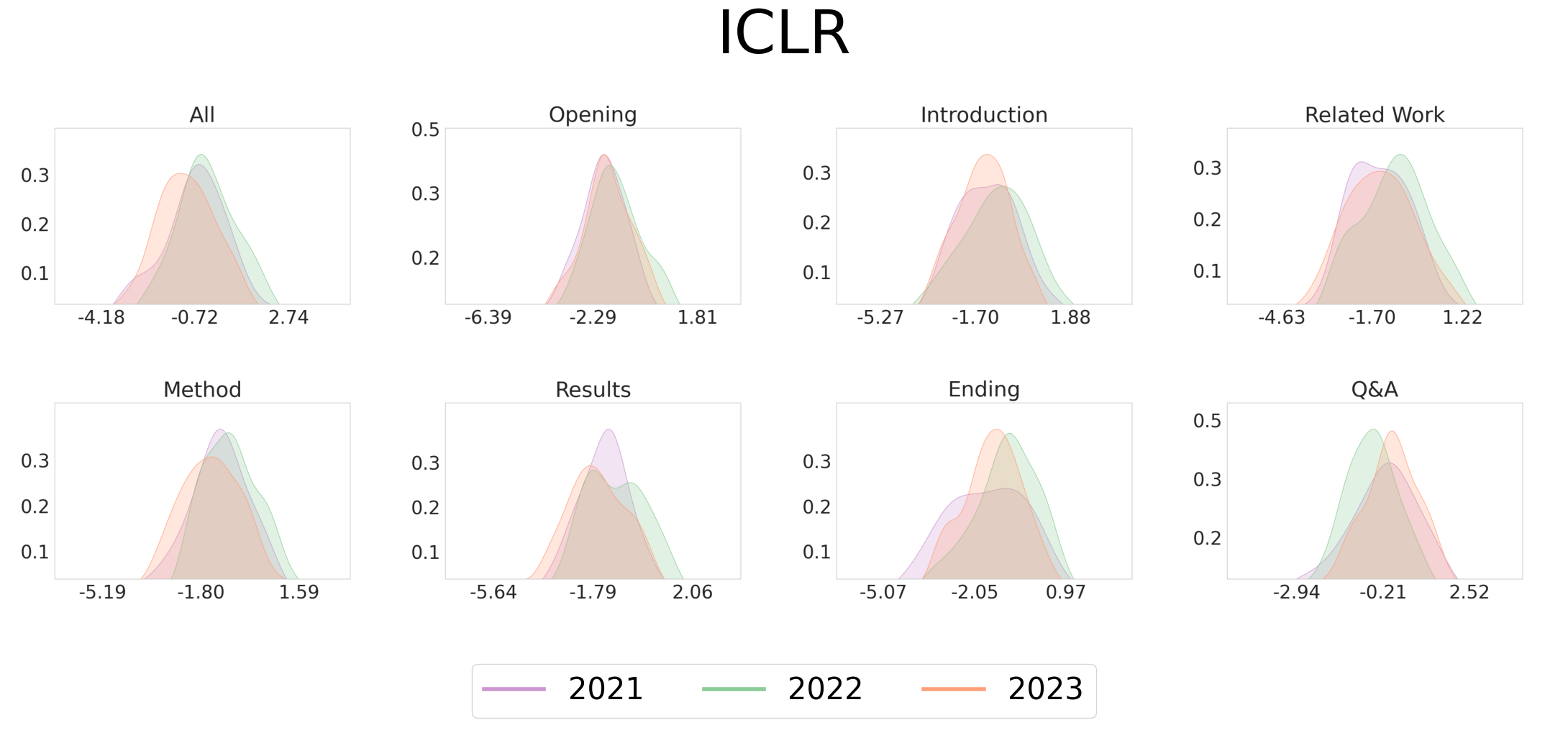}
    \caption{Criterion distribution in ICLR oral presentations.}
    \label{MGT_ICLR}
\end{figure*}

\begin{figure*}[t]
    \centering
    \includegraphics[width=\textwidth]{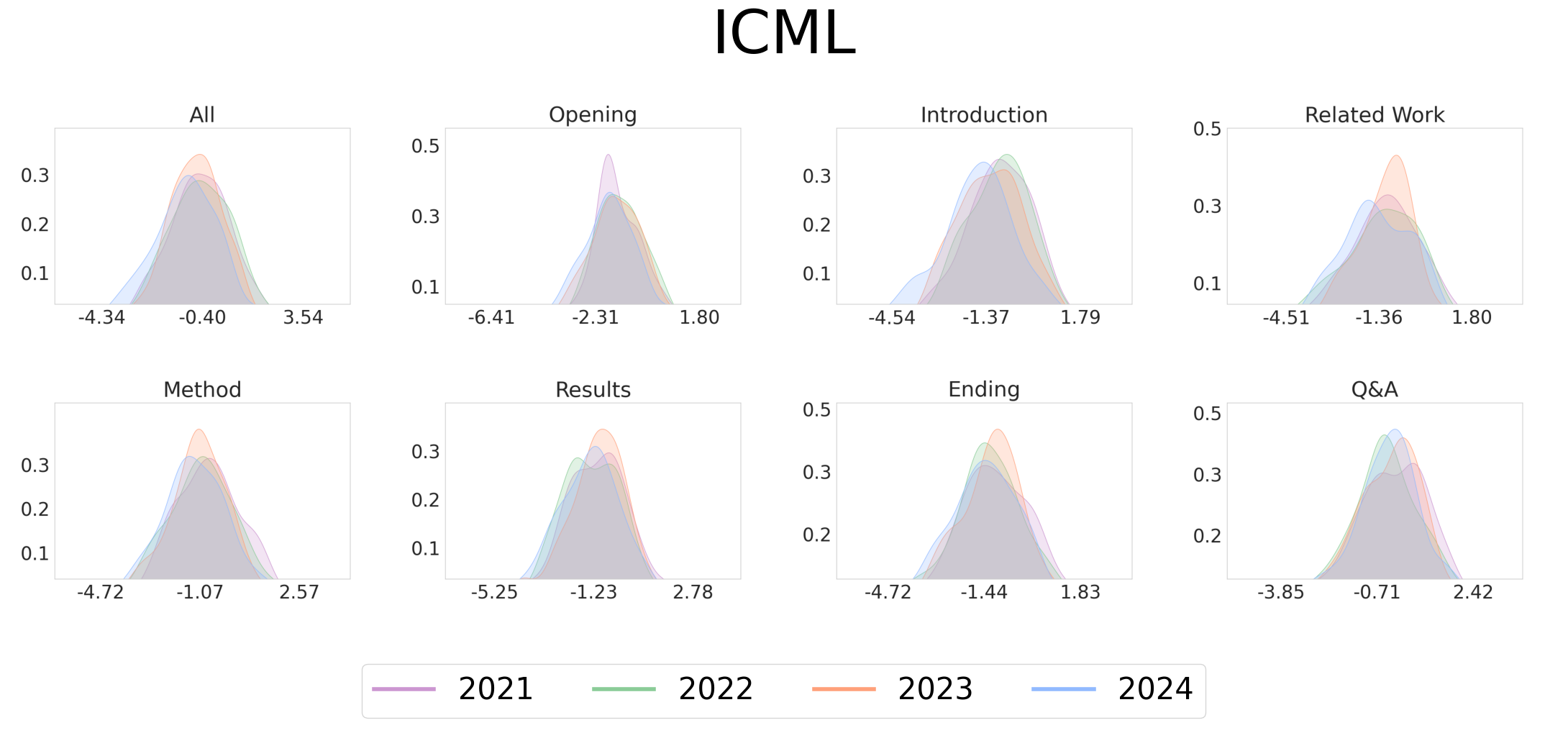}
    \caption{Criterion distribution in ICML oral presentations.}
    \label{MGT_ICML}
\end{figure*}

\begin{figure*}[t]
    \centering
    \includegraphics[width=\textwidth]{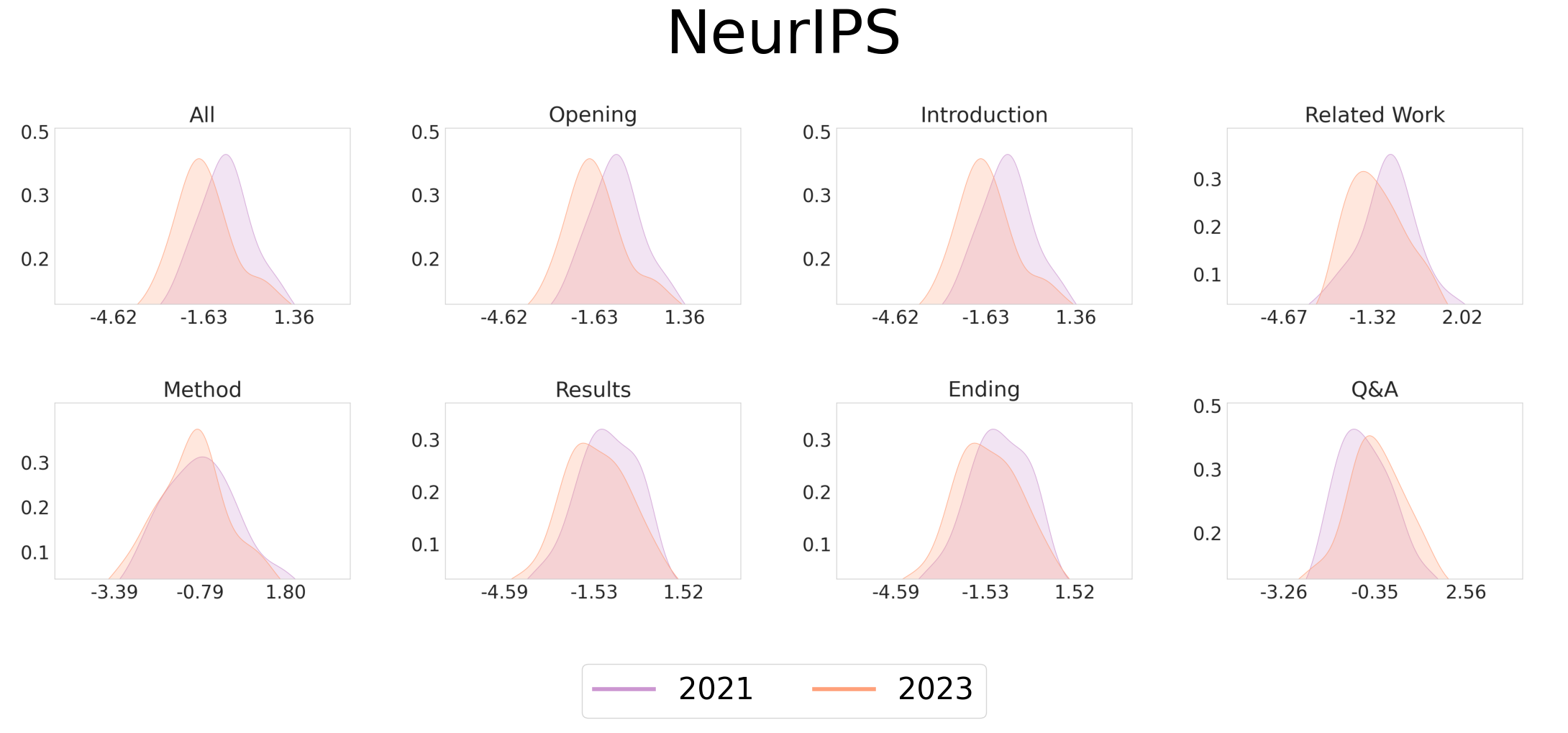}
    \caption{Criterion distribution in NeurIPS oral presentations.}
    \label{MGT_NeurIPS}
\end{figure*}

\begin{figure*}[t]
    \centering 
    \includegraphics[width=\textwidth]{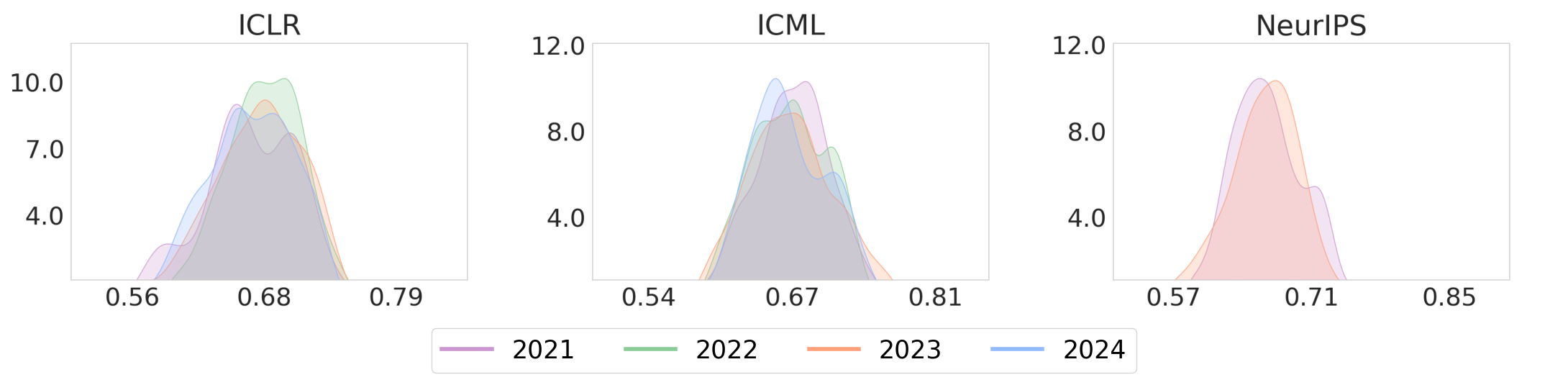}
    \caption{Green class of GLTR in abstract of oral works.}
    \label{gltr-abstract-green}
\end{figure*}

\begin{figure*}[t]
    \centering
    \includegraphics[width=\textwidth]{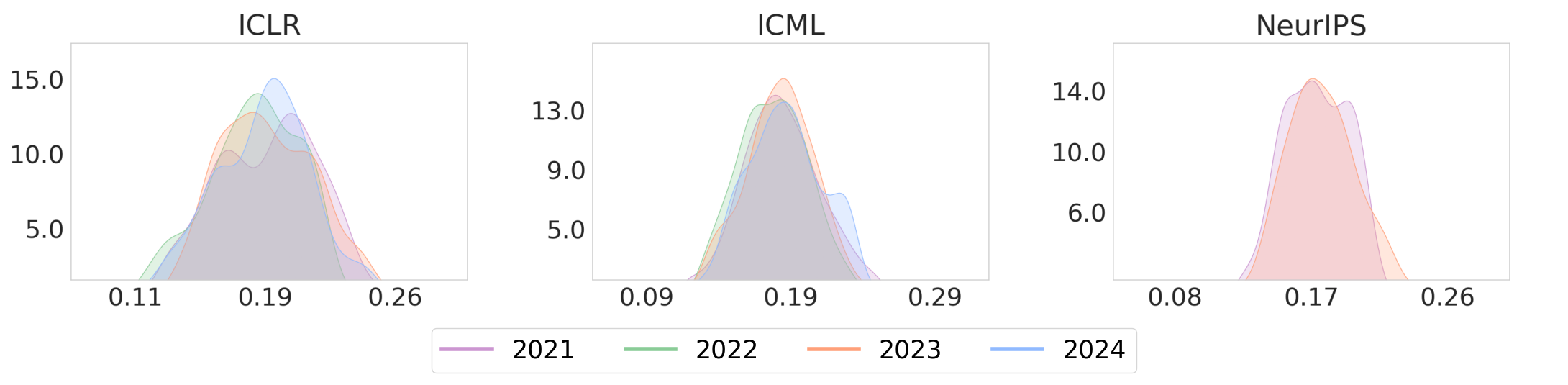}
    \caption{Yellow class of GLTR in abstract of oral works.}
    \label{gltr-abstract-yellow}
\end{figure*}

\begin{figure*}[t]
    \centering
    \includegraphics[width=\textwidth]{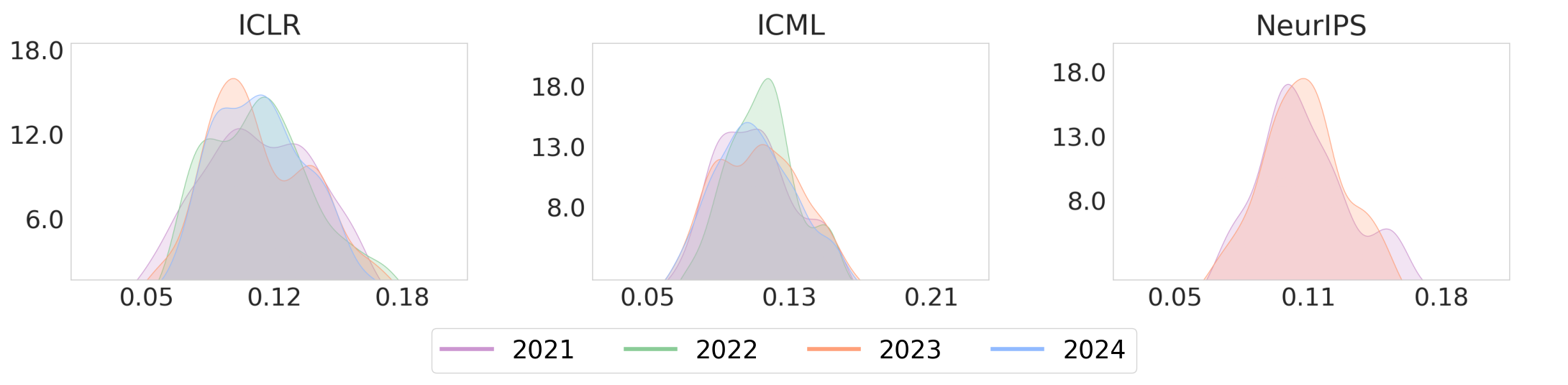}
    \caption{Red class of GLTR in abstract of oral works.}
    \label{gltr-abstract-red}
\end{figure*}

\begin{figure*}[t]
    \centering
    \includegraphics[width=\textwidth]{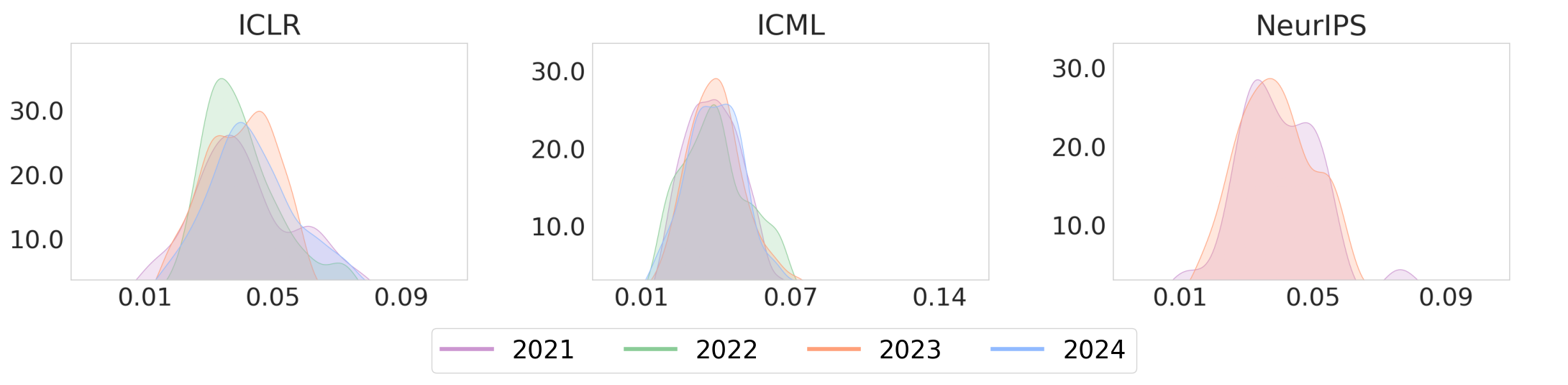}
    \caption{Purple class of GLTR in abstract of oral works.}
    \label{gltr-abstract-purple}
\end{figure*}

\begin{figure*}[t]
    \centering
    \includegraphics[width=\textwidth]{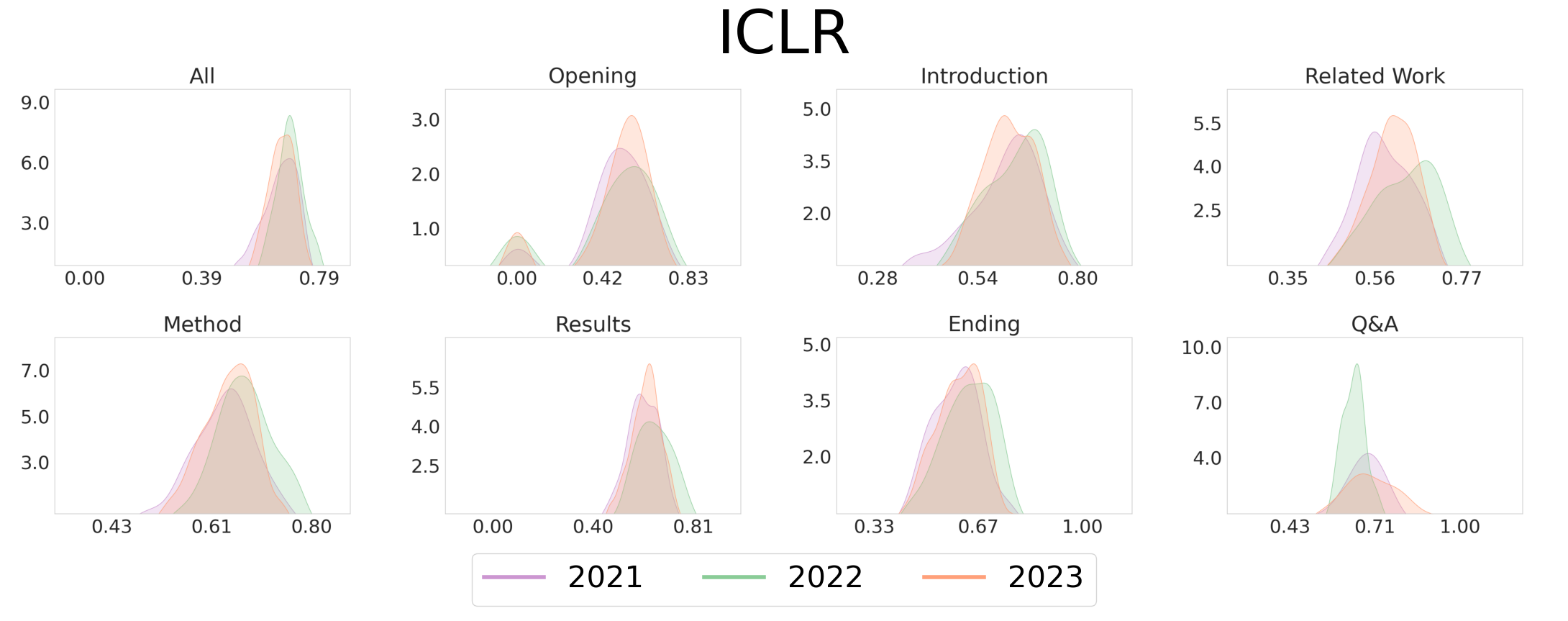}
    \caption{Green class of GLTR in ICLR oral presentations.}
    \label{gltr-ICLR-green}
\end{figure*}

\begin{figure*}[t]
    \centering
    \includegraphics[width=\textwidth]{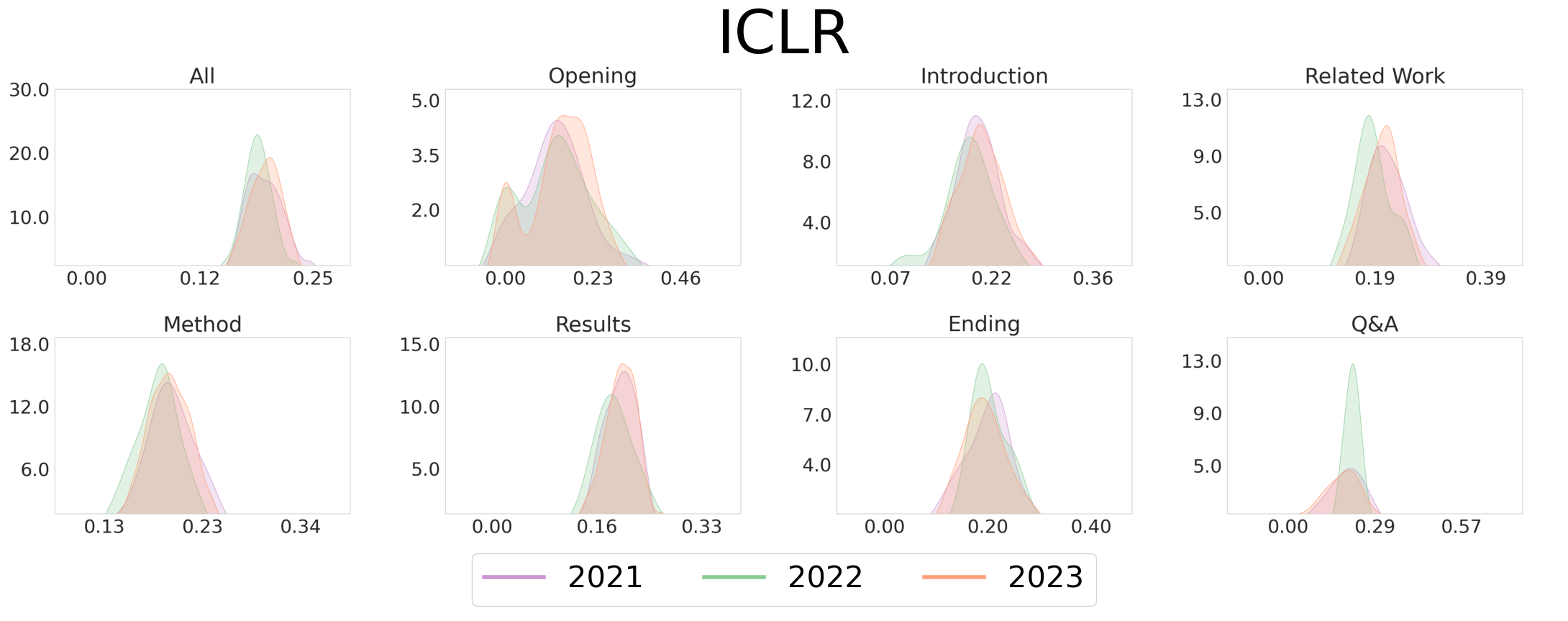}
    \caption{Yellow class of GLTR in ICLR oral presentations.}
    \label{gltr-ICLR-yellow}
\end{figure*}

\begin{figure*}[t]
    \centering
    \includegraphics[width=\textwidth]{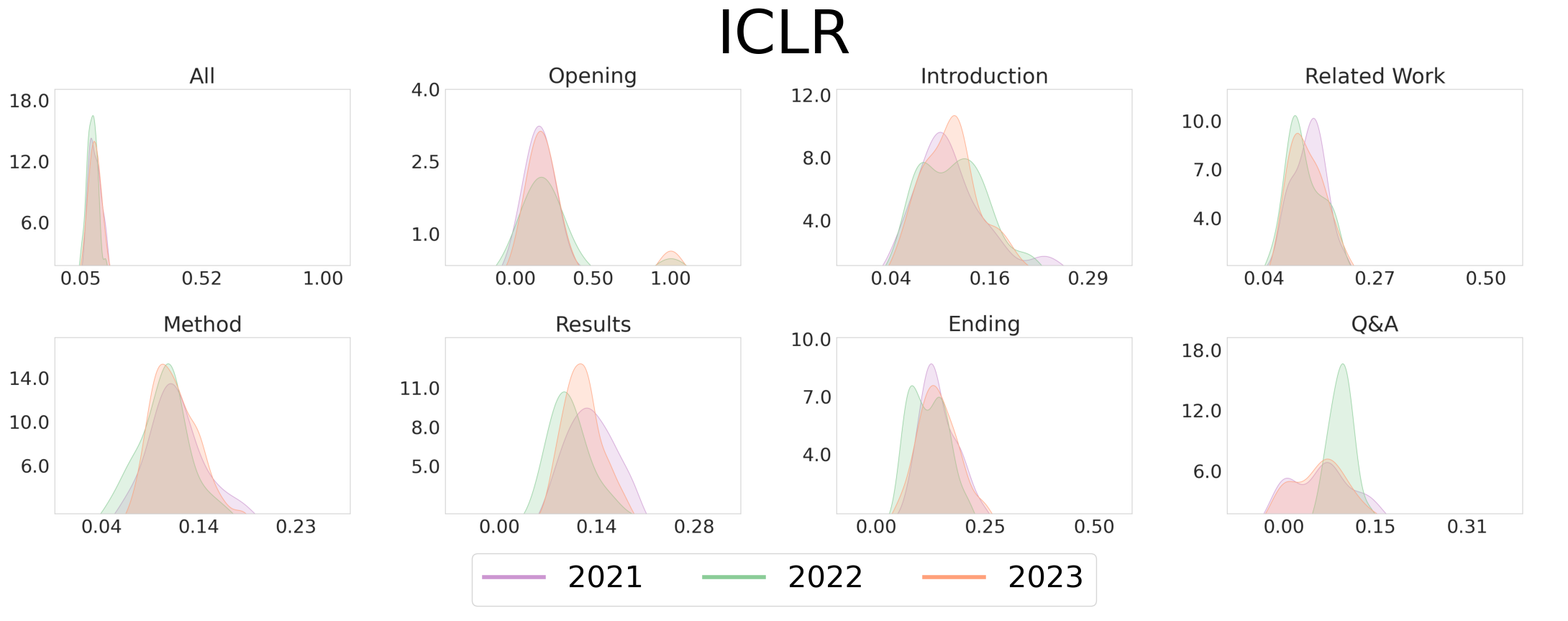}
    \caption{Red class of GLTR in ICLR oral presentations.}
    \label{gltr-ICLR-red}
\end{figure*}

\begin{figure*}[t]
    \centering
    \includegraphics[width=\textwidth]{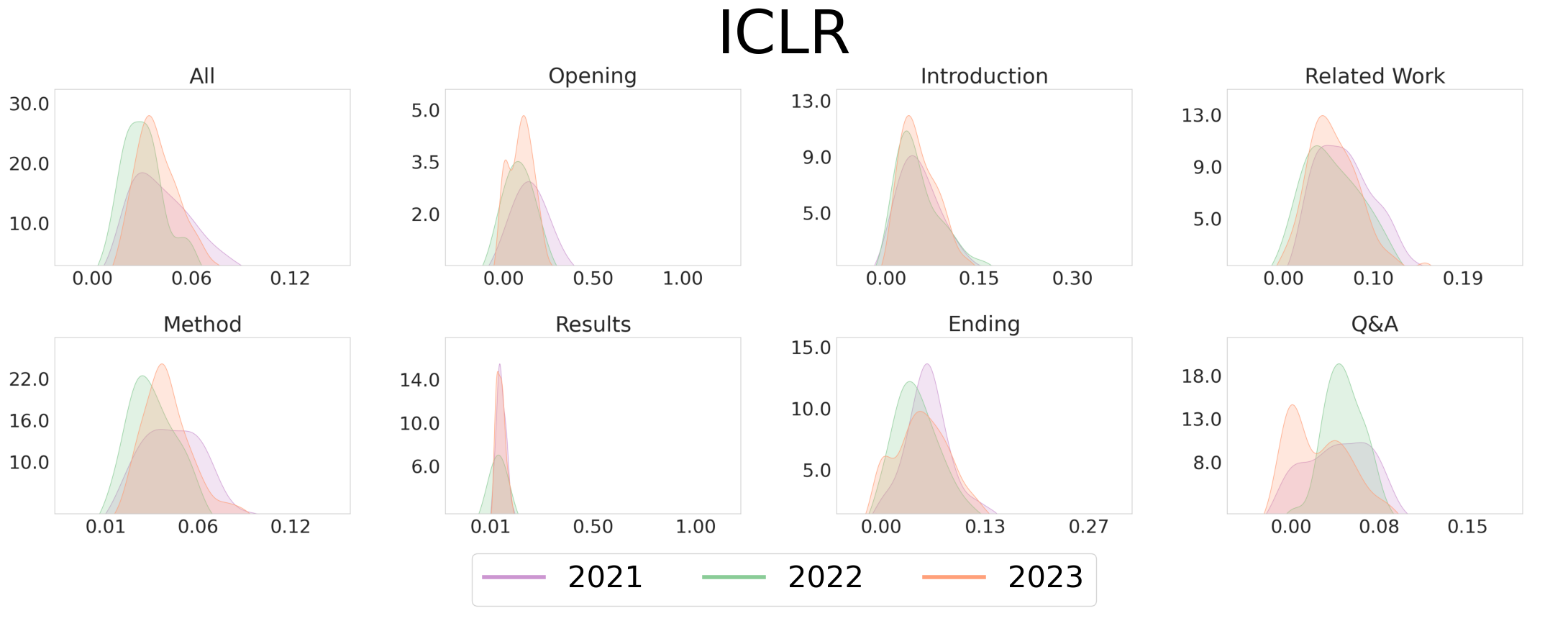}
    \caption{Purple class of GLTR in ICLR oral presentations.}
    \label{gltr-ICLR-purple}
\end{figure*}

\begin{figure*}[t]
    \centering
    \includegraphics[width=\textwidth]{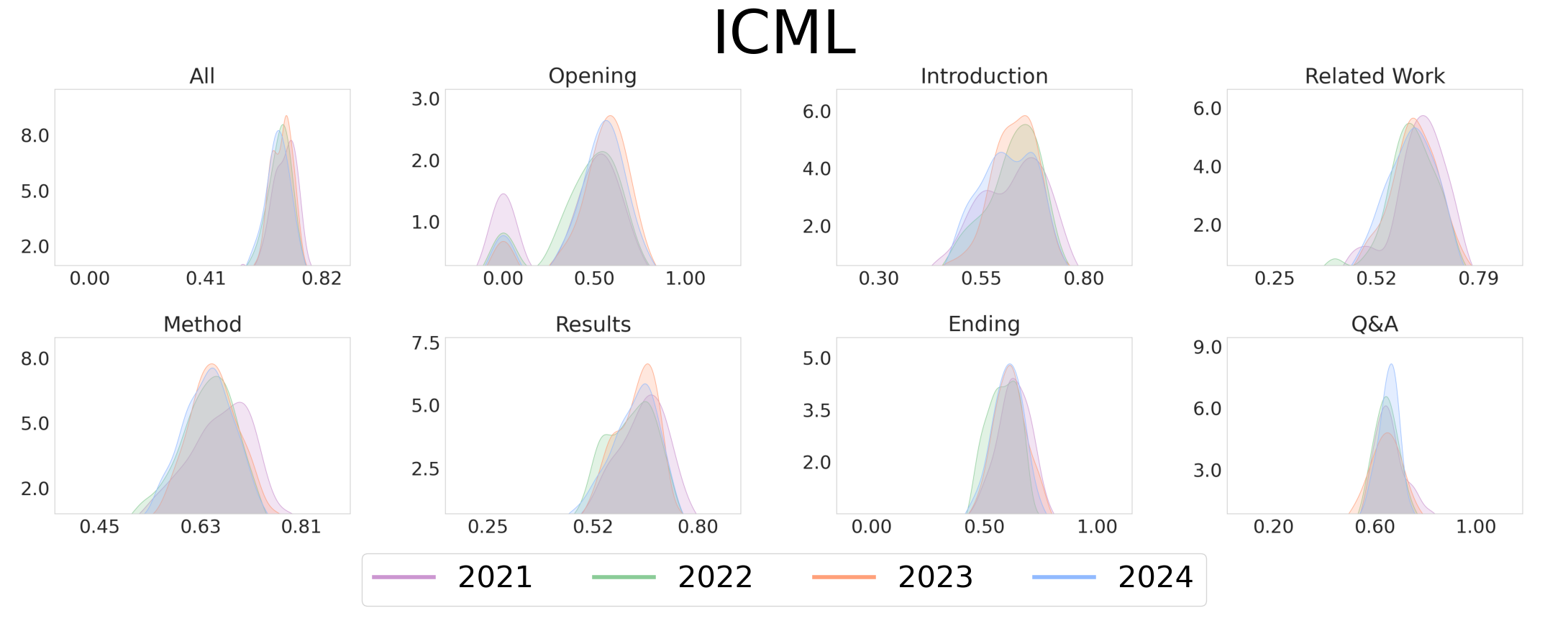}
    \caption{Green class of GLTR in ICML oral presentations.}
    \label{gltr-ICML-green}
\end{figure*}

\begin{figure*}[t]
    \centering
    \includegraphics[width=\textwidth]{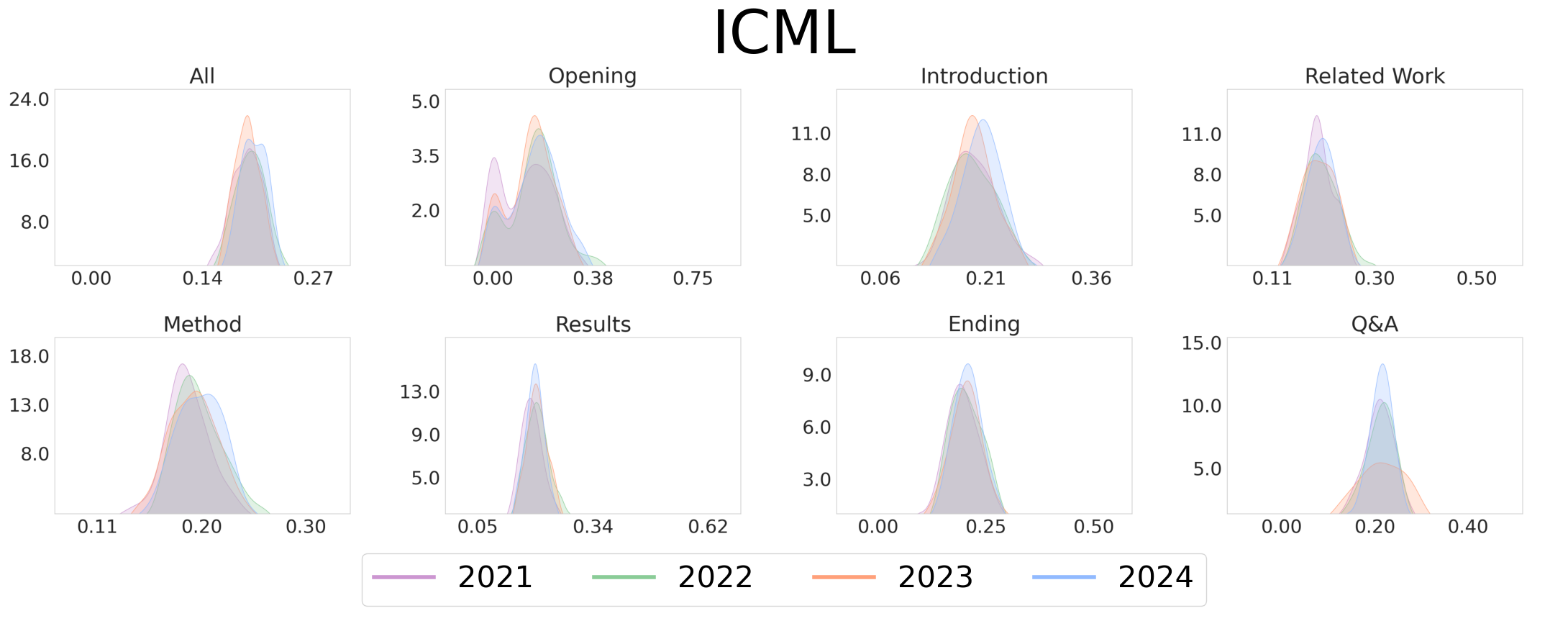}
    \caption{Yellow class of GLTR in ICML oral presentations.}
    \label{gltr-ICML-yellow}
\end{figure*}

\begin{figure*}[t]
    \centering
    \includegraphics[width=\textwidth]{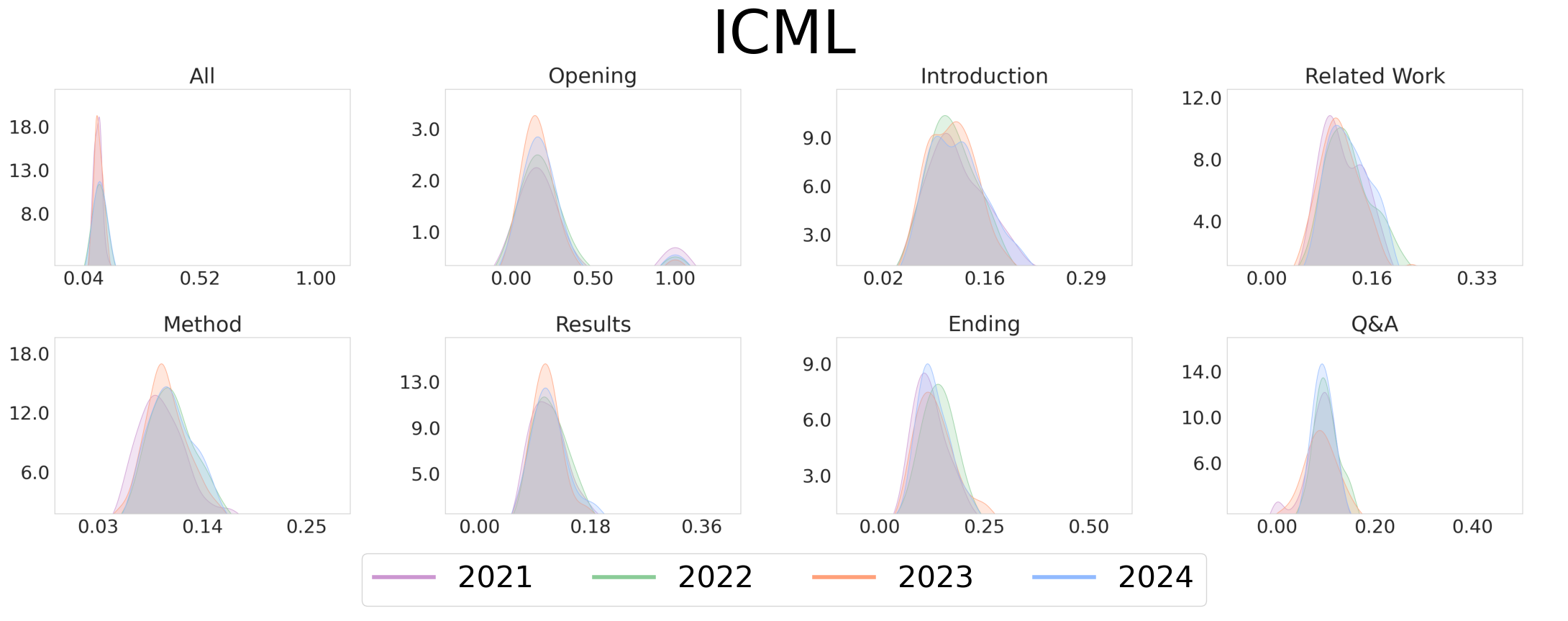}
    \caption{Red class of GLTR in ICML oral presentations.}
    \label{gltr-ICML-red}
\end{figure*}

\begin{figure*}[t]
    \centering
    \includegraphics[width=\textwidth]{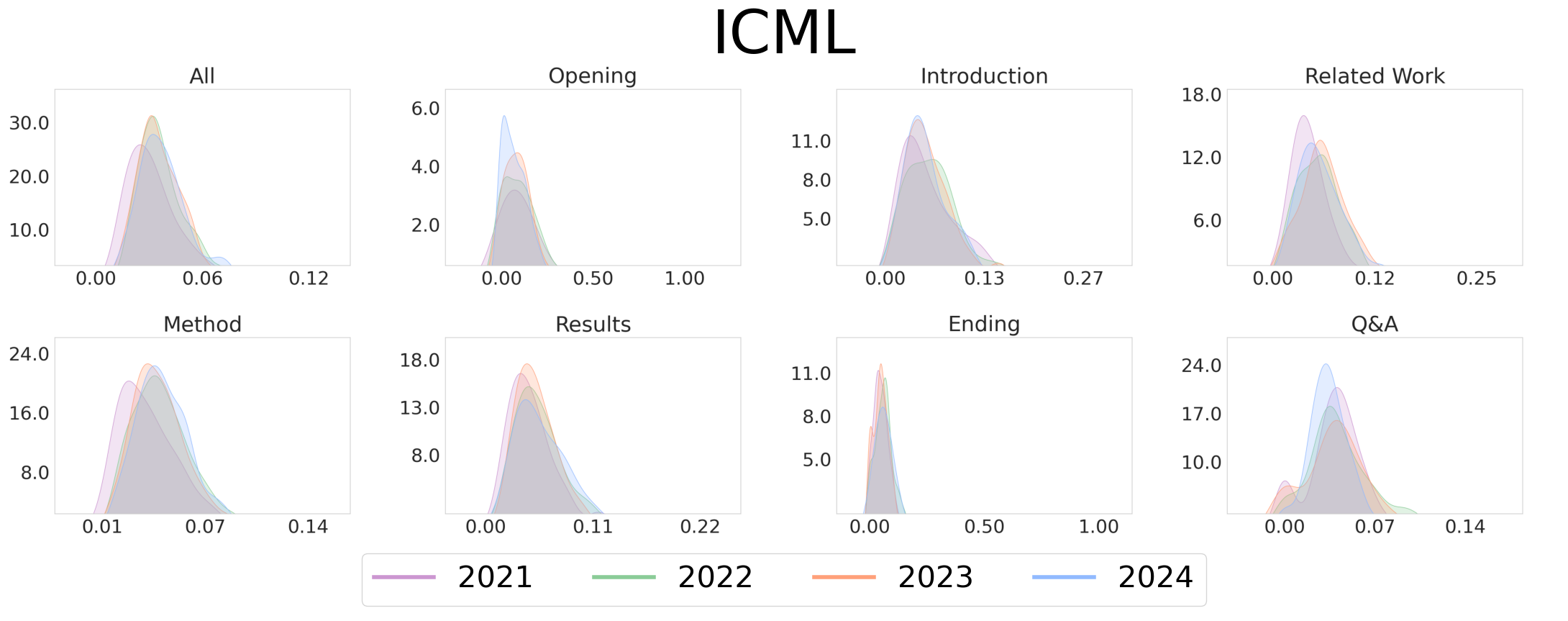}
    \caption{Purple class of GLTR in ICML oral presentations.}
    \label{gltr-ICML-purple}
\end{figure*}

\begin{figure*}[t]
    \centering
    \includegraphics[width=\textwidth]{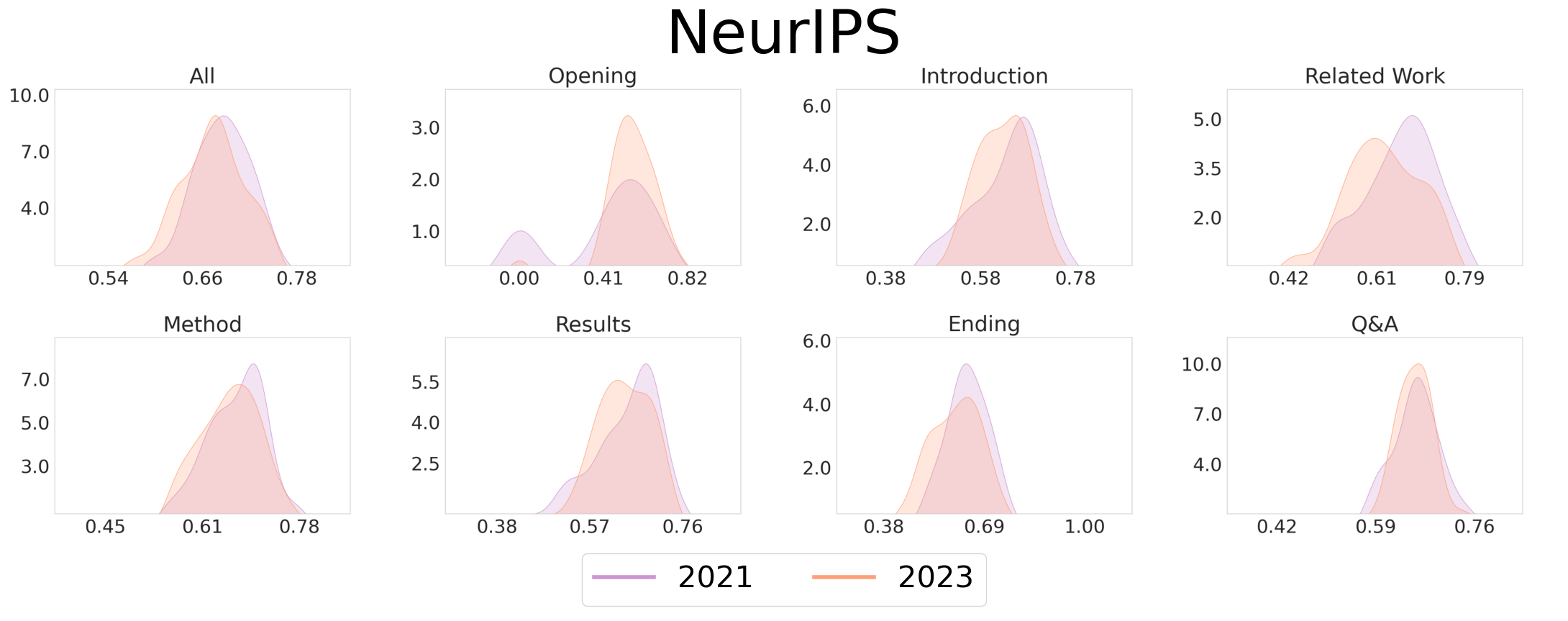}
    \caption{Green class of GLTR in NeurIPS oral presentations.}
    \label{gltr-Nips-green}
\end{figure*}

\begin{figure*}[t]
    \centering
    \includegraphics[width=\textwidth]{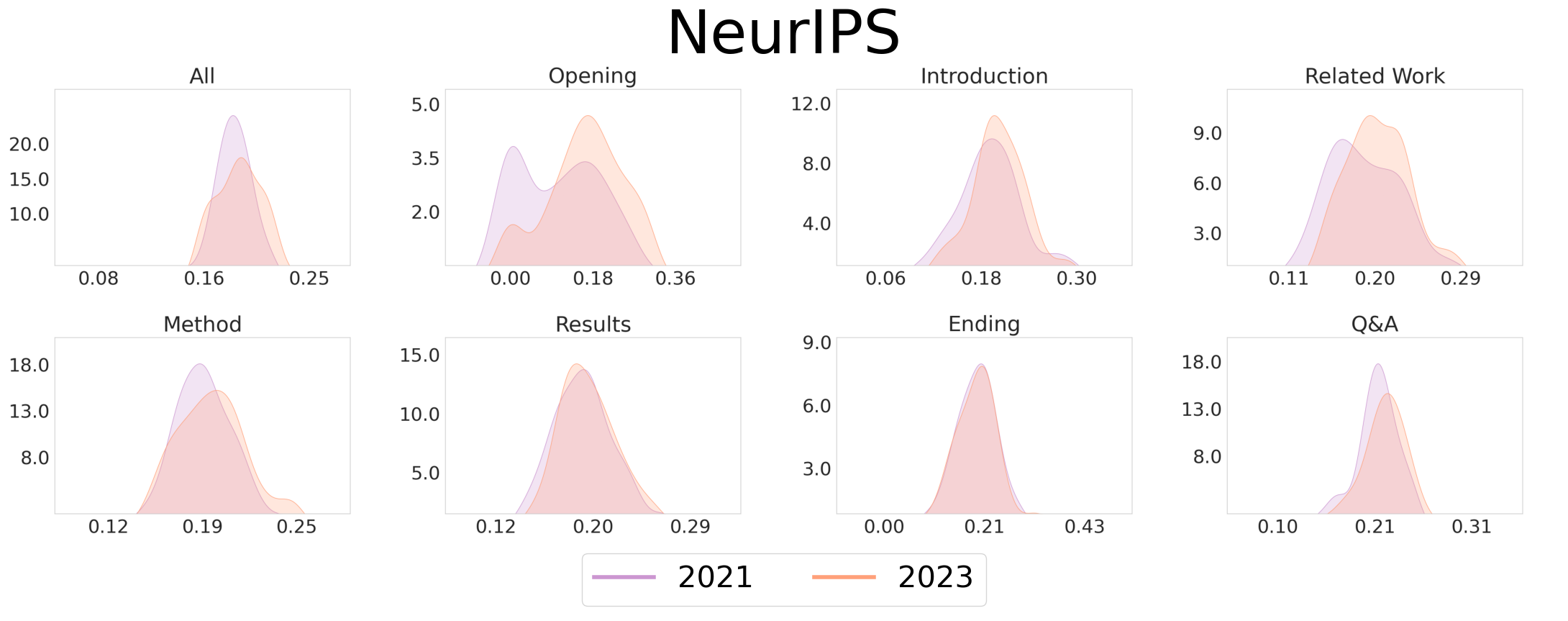}
    \caption{Yellow class of GLTR in NeurIPS oral presentations.}
    \label{gltr-Nips-yellow}
\end{figure*}

\begin{figure*}[t]
    \centering
    \includegraphics[width=\textwidth]{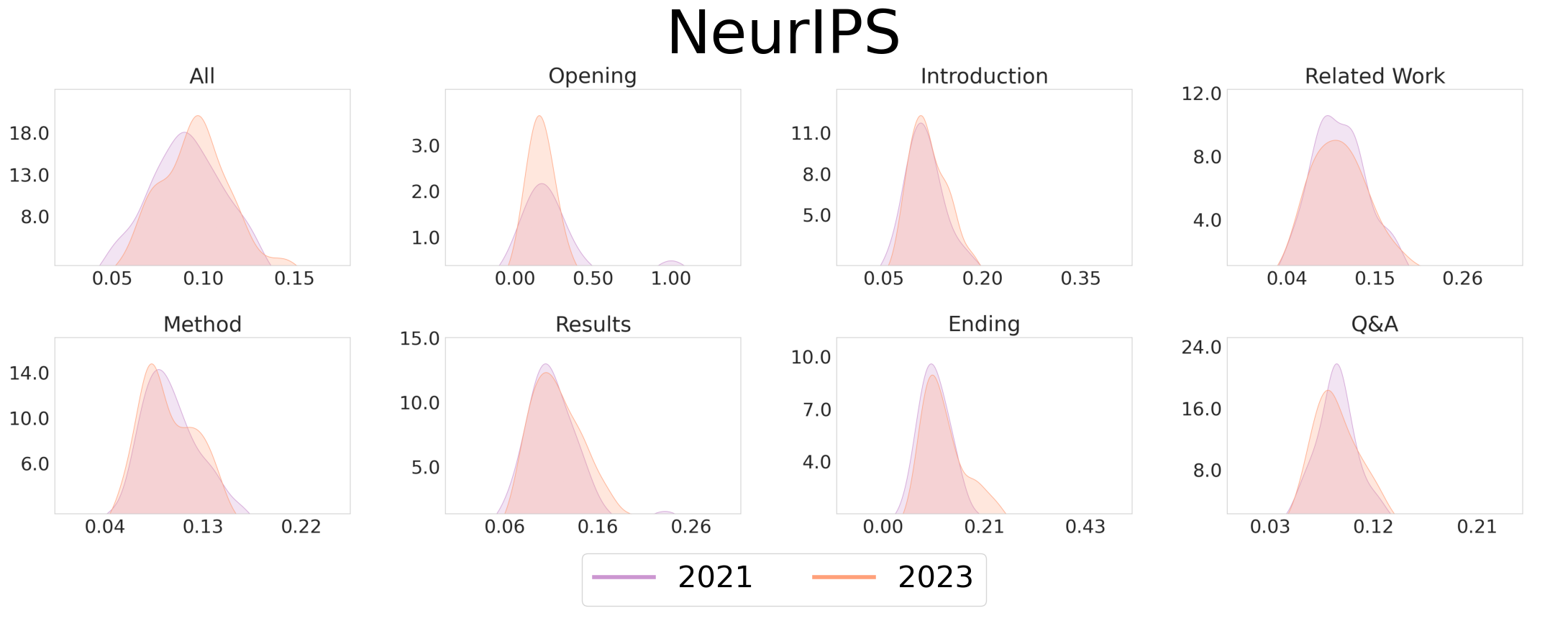}
    \caption{Red class of GLTR in NeurIPS oral presentations.}
    \label{gltr-Nips-red}
\end{figure*}

\begin{figure*}[t]
    \centering
    \includegraphics[width=\textwidth]{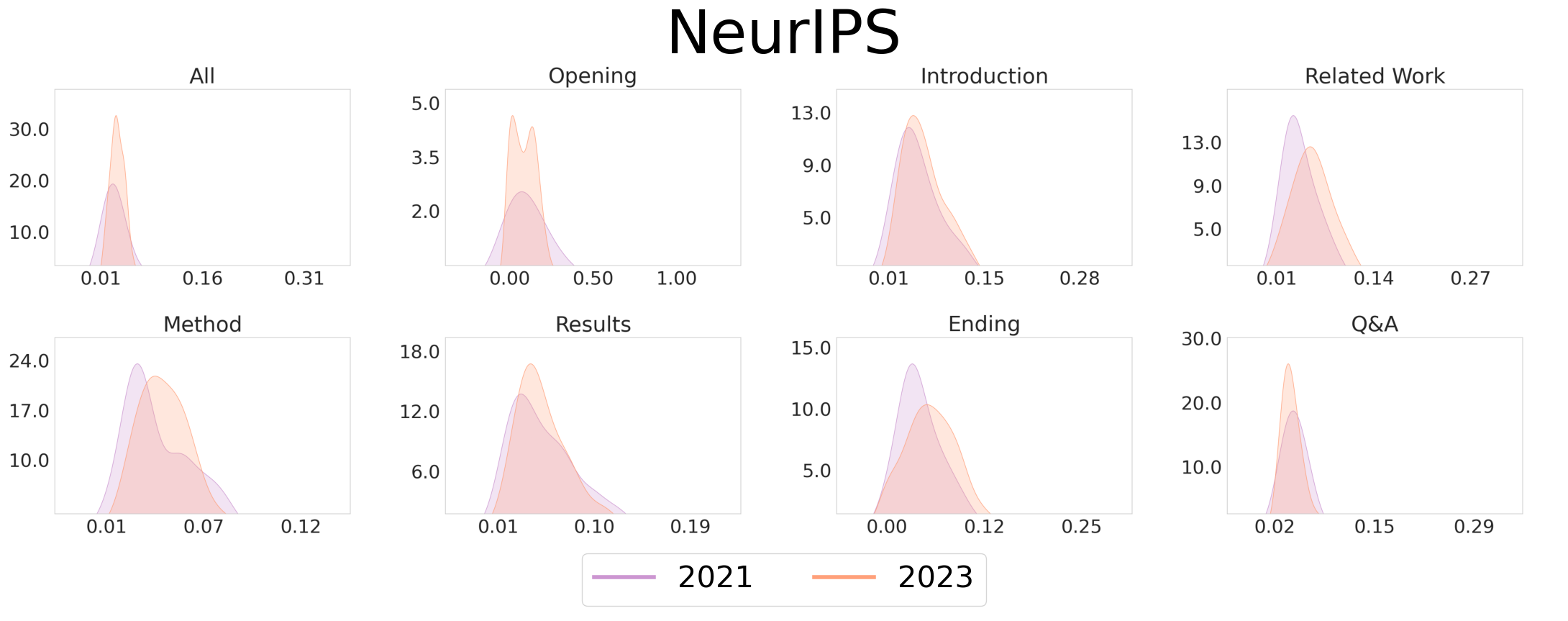}
    \caption{Purple class of GLTR in NeurIPS oral presentations.}
    \label{gltr-Nips-purple}
\end{figure*}

\begin{figure*}[t]
    \centering
    \includegraphics[width=\textwidth]{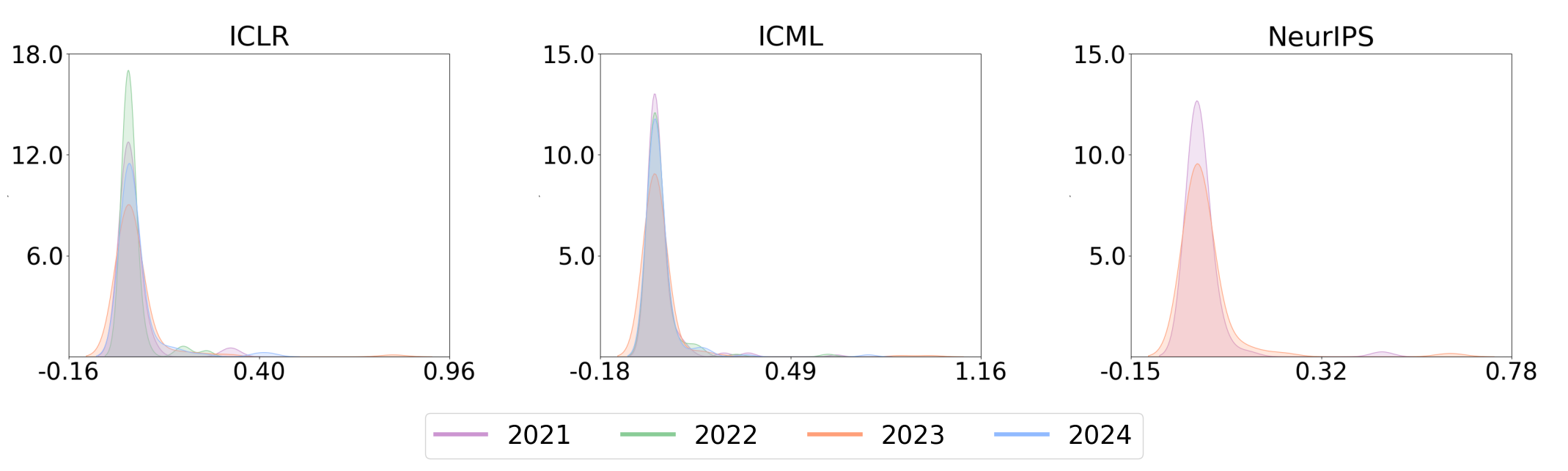}
    \caption{AI-Generated probability in abstracts of oral works.}
    \label{radar-abstract}
\end{figure*}

\begin{figure*}[t]
    \centering
    \includegraphics[width=\textwidth]{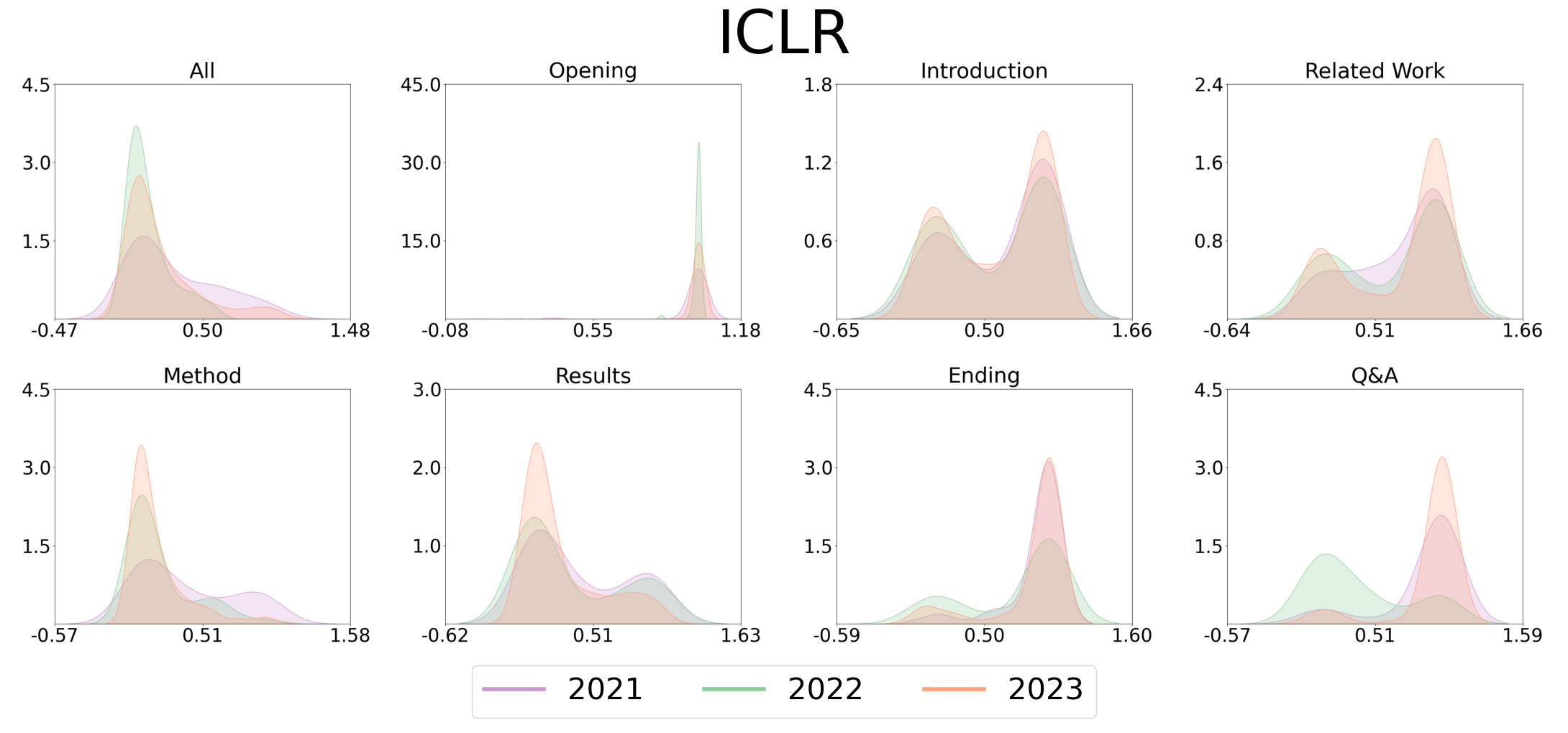}
    \caption{AI-Generated probability in ICLR oral presentations.}
    \label{radar-ICLR}
\end{figure*}

\begin{figure*}[t]
    \centering
    \includegraphics[width=\textwidth]{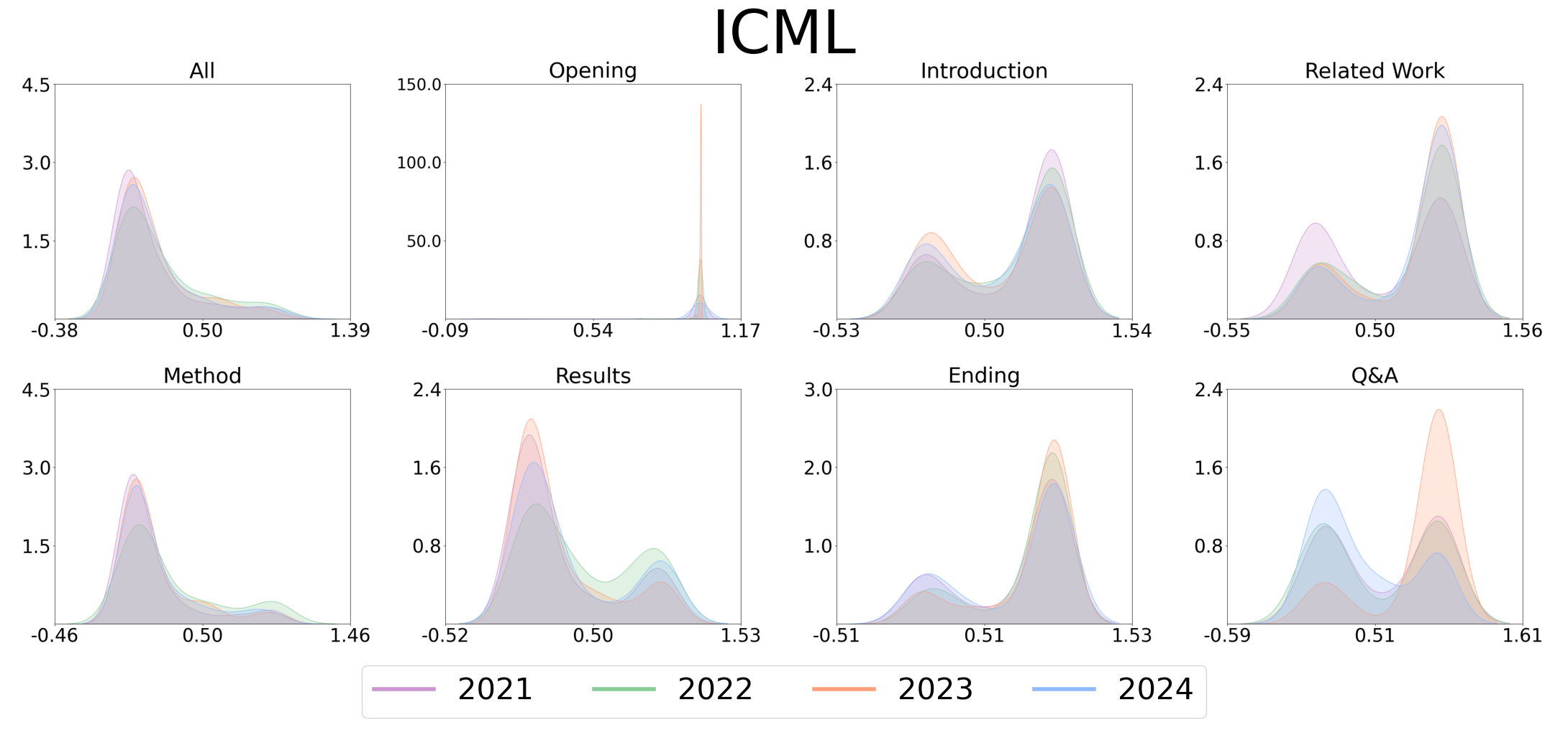}
    \caption{AI-Generated probability in ICML oral presentations.}
    \label{radar-ICML}
\end{figure*}

\begin{figure*}[t]
    \centering
    \includegraphics[width=\textwidth]{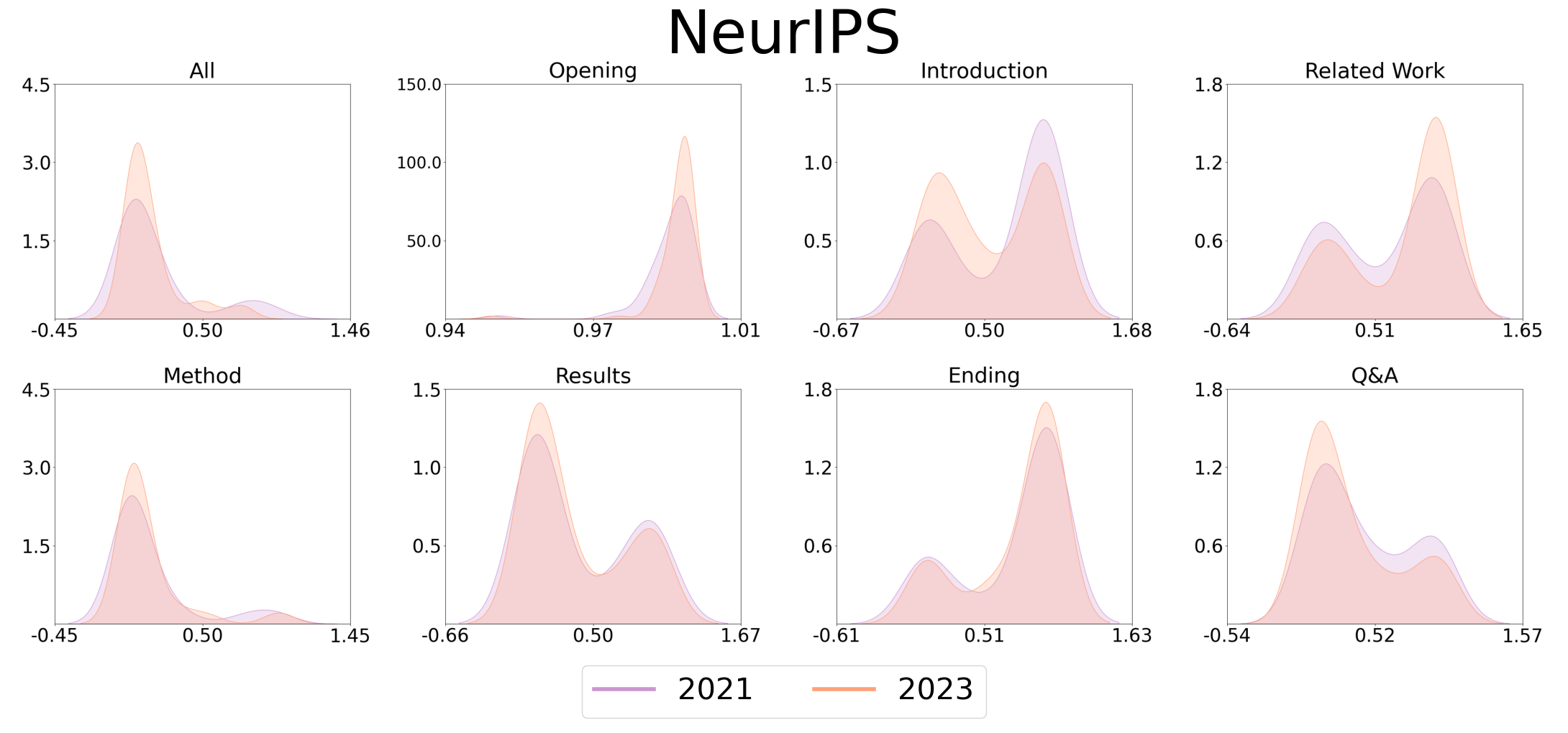}
    \caption{AI-Generated probability in NeurIPS oral presentations.}
    \label{radar-Nips}
\end{figure*}

In this section, we present the detection results from Fast-DetectGPT \citep{bao2023fast}, GLTR \citep{gehrmann2019gltr} and RADAR \citep{Hu2023RADARRA}. All experiments are conducted in a dual-4090 server, detailed as follows:

\begin{itemize}[itemsep=0pt, leftmargin=*]
    \item \textbf{Fast-DetectGPT} serves as a coarse binary-classification, and the results are summarized in Figure \ref{Oral_abstract} and detailed in Figures \ref{MGT_ICLR}, \ref{MGT_ICML}, and \ref{MGT_NeurIPS}. A higher criterion represents a greater probability of machine-generated text.
    \item \textbf{GLTR} works as a coarse four-class classification, and the results are summarized in Figures \ref{gltr-abstract-green} to \ref{gltr-abstract-purple} and detailed in Figures \ref{gltr-ICLR-green} to \ref{gltr-Nips-purple}. It analyzes GPT-2's predictions at each position in the text and calculates the rank of every word. Words ranked in the top 10 are classified as green, those in the top 100 as yellow, in the top 1000 as red, and the rest as purple.
    \item \textbf{RADAR} provides robust binary classification experiments, and the results are summarized in Figure \ref{radar-abstract} and detailed in Figures \ref{radar-ICLR}, \ref{radar-ICML}, and \ref{radar-Nips}. A higher threshold indicates an increased likelihood of identifying machine-generated text through adversarial learning techniques.
\end{itemize}

Consistent with previous analyses, there is no obvious effect of LLMs in the presentations of these oral presentations, demonstrating the need for fine-grained detection such as the word frequency experiments we presented before.

\subsection{Issues in Speech-to-Text and Abstracts}
In this section, we introduce the issues we discover in the speech-to-text process and abstracts, detailed as follows:
\begin{itemize}[itemsep=0pt, leftmargin=*]
    \item There are words such as \textit{``ok''} and \textit{``okay''} which have identical pronunciations and meanings, but Speech-to-Text systems may exhibit a preference for one over the other. For instance, in the ICML oral presentations of 2021 and 2022,\textit{ ``ok''} was predominantly used, whereas in 2023 and 2024, \textit{``okay''} became more common. Such discrepancies introduce inconsistencies that can affect the accuracy of word frequency analyses.
    \item The conversion of technical terms by Speech-to-Text systems may not always be accurate. For example, in multiple oral presentations, the term \textit{``LoRA''} (a well-known PEFT method) was incorrectly transcribed as the name \textit{``Laura''} in the subtitles, which is evidently erroneous. Fortunately, the impact of such errors on the overall word frequency statistics appears to be minimal.
    \item While retrieving abstracts from ICML, ICLR, and NeurIPS conferences, it was observed that certain abstracts contained formatting elements such as italics or citations, e.g., \verb|\textit{approximately valid}| and \verb|\cite{chen2020learning}|. These formatting artifacts can distort the frequency analysis of certain words, such as \textit{``cite.''}
    \item A comparative analysis was conducted between the original subtitles of the oral presentations and the transcripts generated by Whisper's base model \cite{Radford2022RobustSR} from the audio files of the same presentations. This comparison, evaluating accuracy, consistency, and coherence, indicated that the original subtitles were significantly superior to the transcripts generated by Whisper. Therefore, the original subtitles are deemed more appropriate for use in word frequency analysis.
\end{itemize}

\subsection{Word frequency}
\begin{figure}[t]
    \centering
    \begin{subfigure}[b]{0.45\textwidth}
        \includegraphics[width=\textwidth]{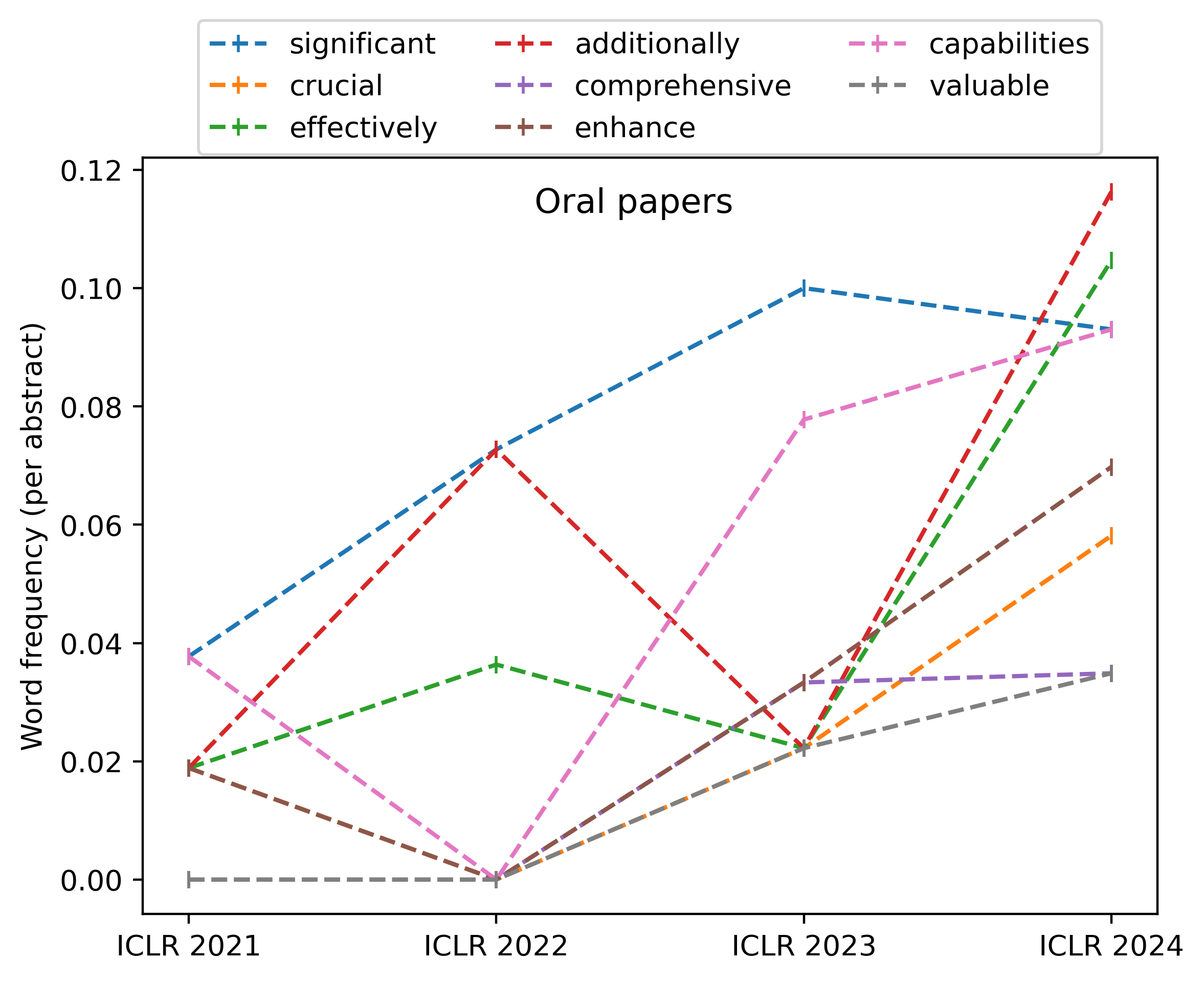}
    \end{subfigure}
    \hfill
    \begin{subfigure}[b]{0.45\textwidth}
        \includegraphics[width=\textwidth]{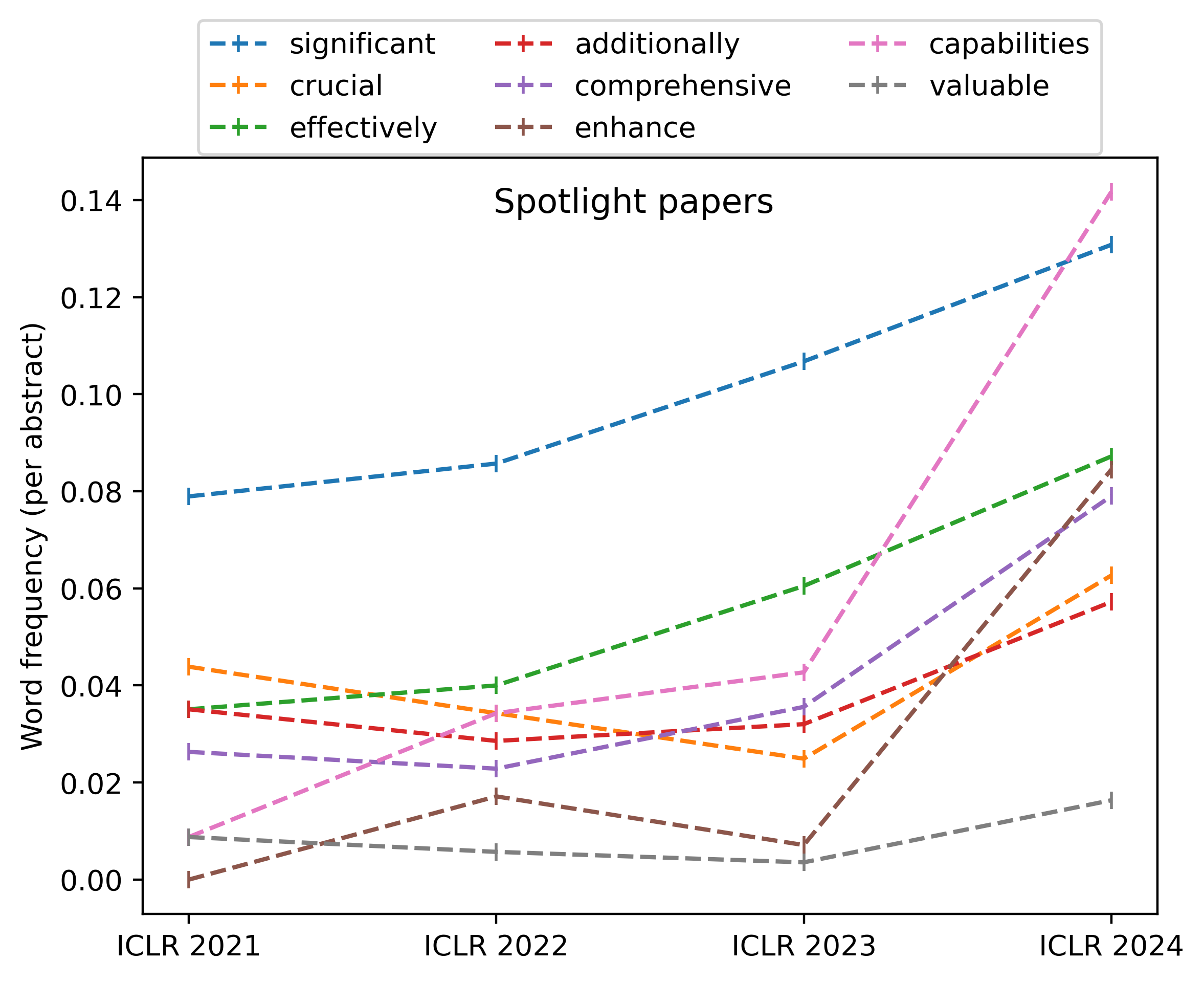}
    \end{subfigure}
    \caption{Word frequency in abstracts for \textit{oral} papers and \textit{spotlight} papers in ICLR.}
    \label{wc_iclr_supp}
\end{figure}

Figure \ref{wc_iclr_supp} also shows the increasing trend in the frequency of these words.

\begin{figure}[t]
    \centering
    \begin{subfigure}[b]{0.45\textwidth}
        \includegraphics[width=\textwidth]{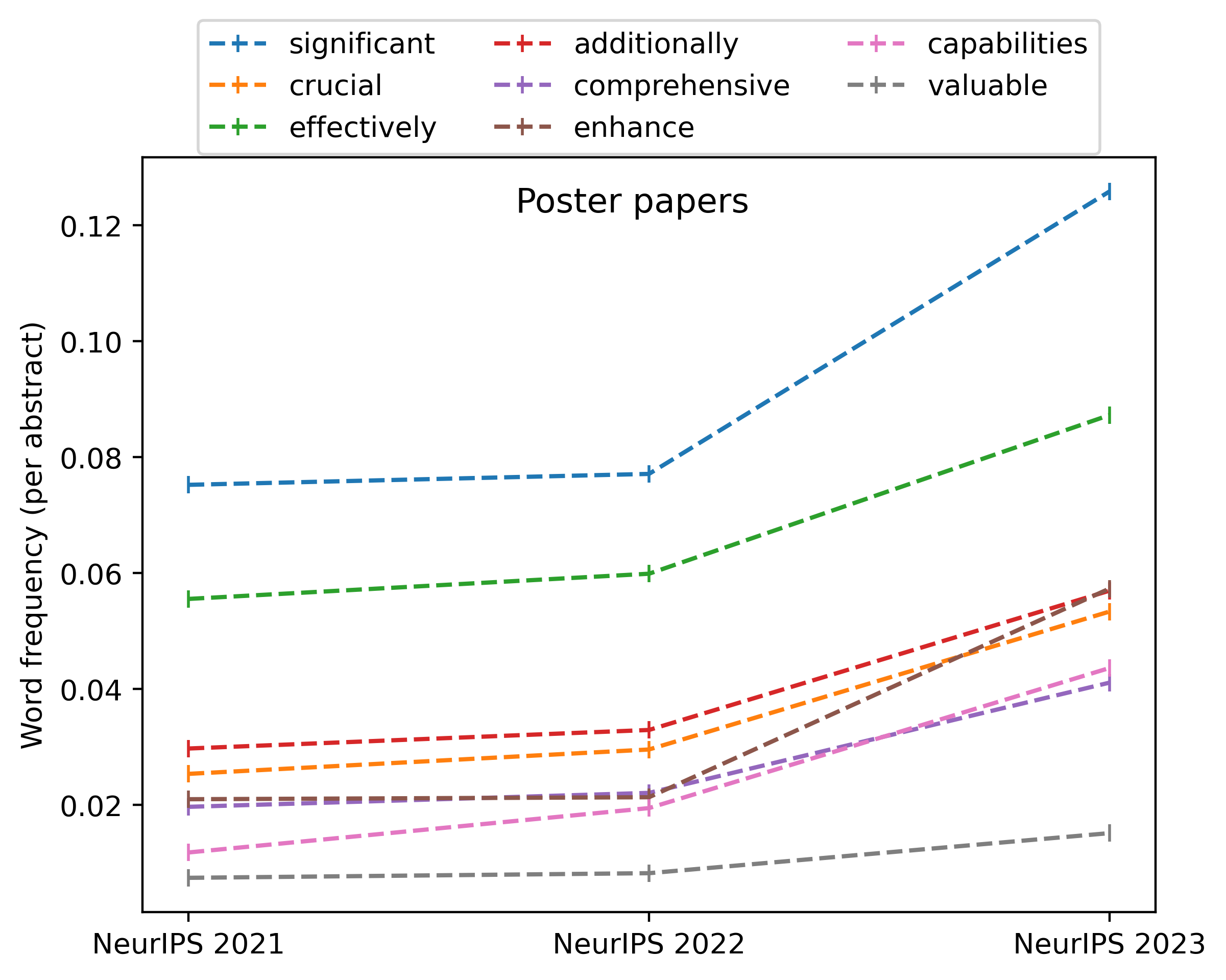}
        \caption{Word frequency in abstracts of \textit{poster} papers.}
        \label{wc_neurips_poster}
    \end{subfigure}
    \hfill
    \begin{subfigure}[b]{0.45\textwidth}
        \includegraphics[width=\textwidth]{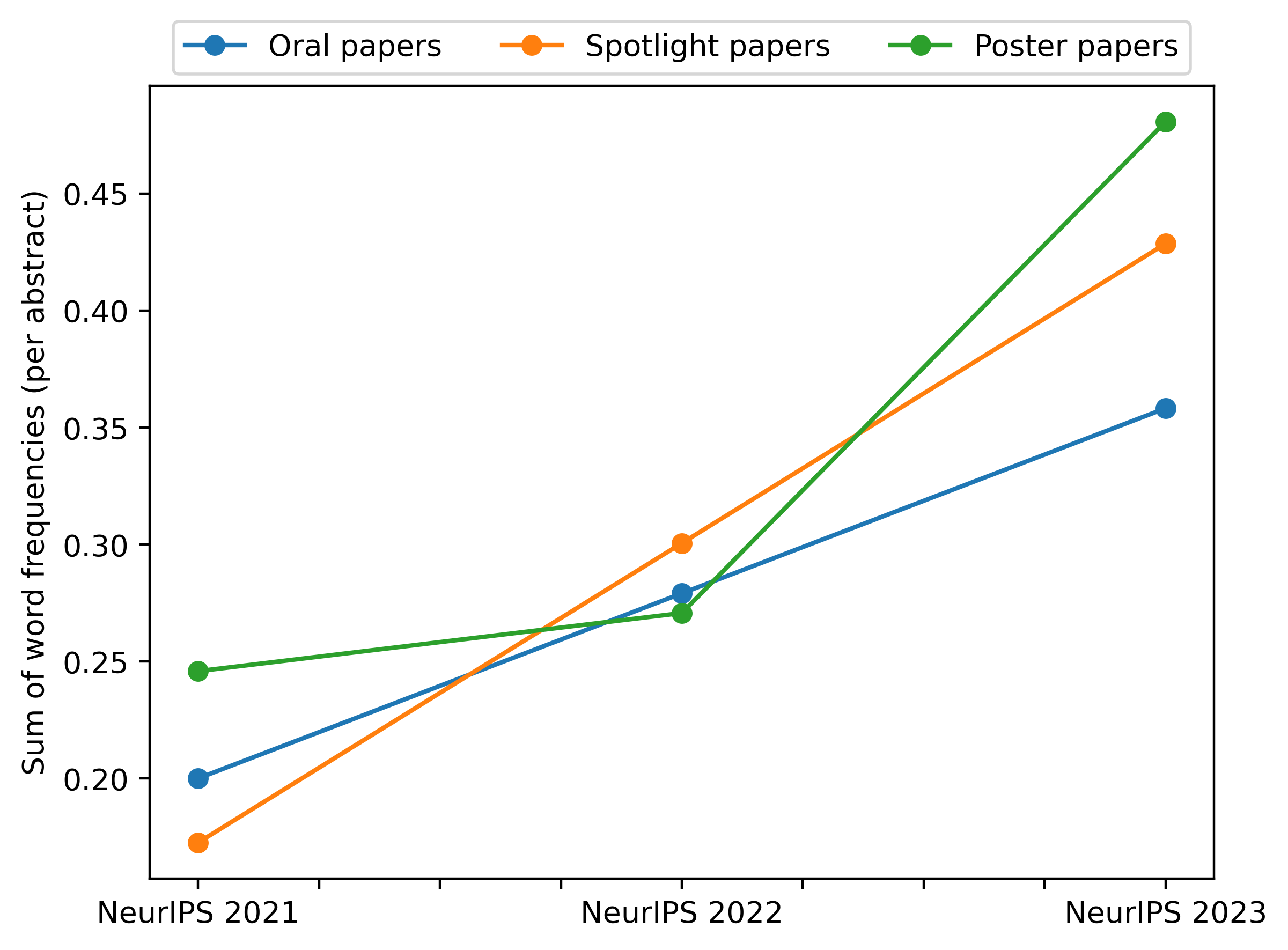}
        \caption{The sum of the frequencies of example words in the abstracts of different types of papers.}
        \label{wc_neurips}
    \end{subfigure}
    \caption{Word frequency in abstracts of NeurIPS papers. Papers accepted in 2022 are considered poster papers, and word frequencies for spotlight and oral papers were completed by interpolation.}
    \label{wc_neurips_all}
\end{figure}

\begin{figure*}[t]
    \centering
    \includegraphics[width=\textwidth]{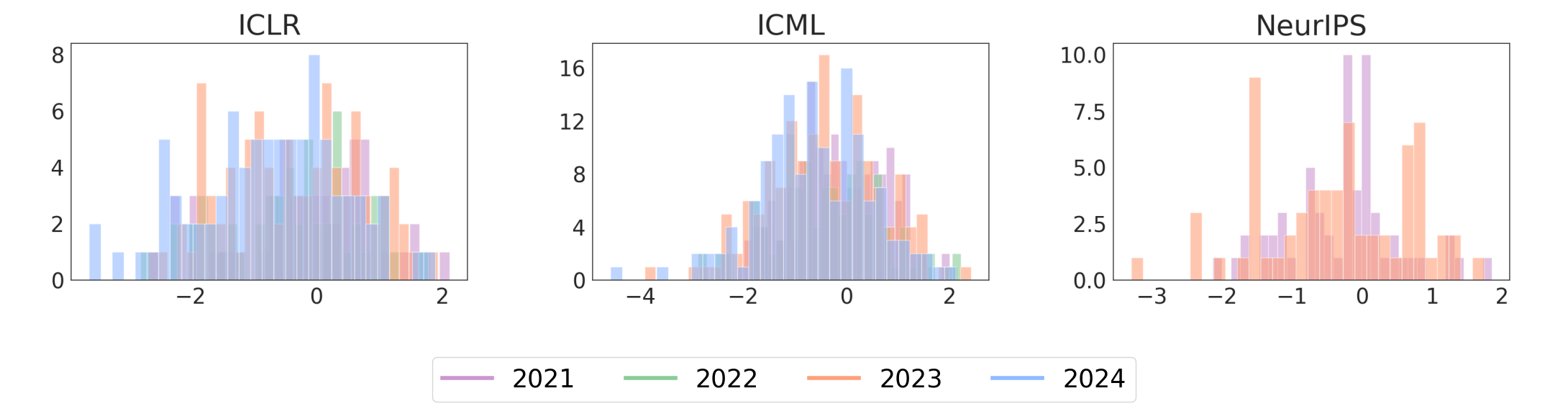}
    \caption{Criterion distribution in abstracts of oral works.}
    \label{Oral_abstract_histogram} 
\end{figure*}

\section{Case Study}
In this section, we sample some oral presentations at each machine learning conference and visualize the MGT detection results. An example of the abstract is shown in Figure \ref{abstract} and the presentation of an oral paper in ICML 2022 shown in Figure \ref{oral-1}, \ref{oral-2}, and \ref{oral-3}.
\begin{figure*}
\begin{tcolorbox}[enhanced,breakable,attach boxed title to top center={yshift=-3mm,yshifttext=-1mm},boxrule=0.9pt, breakable, colback=gray!00,colframe=black!50,colbacktitle=ProcessBlue!70, title=Abstract Example, boxed title style={size=small,colframe=ProcessBlue!70}]
{\small
 Large language models are trained on vast amounts of internet data, prompting concerns that they have memorized public benchmarks. Detecting this type of contamination is challenging because the pretraining data used by proprietary models are often not publicly accessible.\\
 We propose a procedure for detecting test set contamination of language models with exact false positive guarantees and without access to pretraining data or model weights. Our approach leverages the fact that when there is no data contamination, all orderings of an exchangeable benchmark should be equally likely. In contrast, the tendency for language models to memorize example order means that a contaminated language model will find certain canonical orderings to be much more likely than others. Our test flags potential contamination whenever the likelihood of a canonically ordered benchmark dataset is significantly higher than the likelihood after shuffling the examples.\\
 We demonstrate that our procedure is sensitive enough to reliably detect contamination in challenging situations, including models as small as 1.4 billion parameters, on small test sets with only 1000 examples, and datasets that appear only a few times in the pretraining corpus. Finally, we evaluate LLaMA-2 to apply our test in a realistic setting and find our results to be consistent with existing contamination evaluations.}
\end{tcolorbox}
\caption{Abstract Example.}
\label{abstract}
\end{figure*}

\begin{figure*}
\begin{tcolorbox}[enhanced,attach boxed title to top center={yshift=-3mm,yshifttext=-1mm},boxrule=0.9pt,colback=gray!00,colframe=black!50,colbacktitle=ProcessBlue!70,title=Presentation Example - 1,boxed title style={size=small,colframe=ProcessBlue!70},sharp corners,left=5mm,right=5mm,top=2mm,bottom=2mm,boxsep=0pt]
{\small
\hl{'Opening':} 'Hi Embassy Kuala',\\
\hl{'Introduction':} 'And this presentation is about our clear paper titled complex query answering with neuron predictors. And this is joint work with Erik Daniel Helle',\\
\hl{'Related Work':} 'So let's consider the setting. We have a knowledge graph graph structured knowledge base where knowledge about the world is represented in the form of relationships between entities. In the knowledge graph like this, one notes correspond to objects in the world or also upset concepts and I just correspond to relationships between these. In this example we have a knowledge graph about biomedical entities which tells us that paroxetine is used to treat anxiety and as bisimulation as a biological action. And that a pixel band treats deep vein thrombosis and one of its pharmacological effects is that it's an anticoagulant. Now one problem with real world where large scale knowledge graphs is that they are often incomplete. In this particular case, we are missing a link stating the topics abandons a medication that deep vein thrombosis is a disease and that a pixel band kind of side effects when taken together with oxygen. A very effective solution to identifying missing links in large knowledge graphs is via neural in production models in neural prediction. The underlying idea is that we can learn an embedding vector for all the nodes in the graph. For example, in this case we will learn an embedding back to for a pixel band proximity in bisimulation, Deep vein thrombosis and all other entities in the graph. Now assume we want to know the type of relationship if any between a pixel band and boxed in the likelihood that two entities in this case picks abandoned parks 18 are linked by a given type of relationship in this case interacts is a function of the embedding vectors of the source node. In this case a pixel man and embedding vector of the target node in this case, proximity in. And we can use this function for ranking missing links and find out that for example, a pixel bunnies are likely to interact with oxytocin. Even if this link is not directly available in the knowledge graph. Now consider the problem of answering complex queries on incomplete knowledge graphs. Here we have a query which medications have side effects when taken with drugs for treating anxiety and we want to have a list of medications that hands are query. This query can be formalized in logic form and it reads as follows, find M where M is available such that there is a D. Which is also variable, such that I am interacts with the and the tweets anxiety know that this is just an example and our method supports arbitrarily complex logic queries with conjunctions and dysfunctions. Now the best solution for solving this problem proposed so far is the following. First we automatically generate millions of complex query answer path and then we train a deep neural network to produce the correct answers given the creator, the neural model works as follows. First we represent the complex query as a graph. We were each note um corresponds either to a variable or to an entity in the query. Then the graph is passes through a deep neural network which will return ranking list of answers. Now that two main problems with this approach, one is that training is extremely expensive, since the models need to be trained on millions of genetic queries. Also, it's not really clear what happens if we evaluate on queries that differ from the queries that we used for training. Another issue is that there is no explanation for the reasons why a given answer was predicted to solve these problems.'}
\end{tcolorbox}
\caption{Presentation Example - 1}
\label{oral-1}
\end{figure*}

\begin{figure*}
\begin{tcolorbox}[enhanced,attach boxed title to top center={yshift=-3mm,yshifttext=-1mm},boxrule=0.9pt,colback=gray!00,colframe=black!50,colbacktitle=ProcessBlue!70,title=Presentation Example - 2,boxed title style={size=small,colframe=ProcessBlue!70},sharp corners,left=5mm,right=5mm,top=2mm,bottom=2mm,boxsep=0pt]
{\small
\hl{'Method':} 'In this work, we propose a completely new paradigm for answering complex queries on incomplete knowledge graphs. We first train a model F. I, financing simple atomic queries like which drugs to it anxiety. And then we convert its query into an optimization problem where we need to identify the optimal values of the variables in this case, M and D. In the query that maximize the likelihood of both that M interacts with the and likely that the tweets anxiety. Then those two likelihoods are aggregated using a T norm, which is a continuous relaxation of the logical and from fancy logic. And depending on whether we search for the best values of M and the united creator continuous space, we can cast this problem as either as a continuous or discrete optimization problem. In the discrete case, we want to identify the best mapping from variables to entities that maximizes the score of the query. We experimented with a very simple greedy approach to solve this problem, we first start with the variable D. And search for the K most likely values for the by finding the top K. Treatments for anxiety according to the Newell in prediction model. Then for each of the candidate values of the we identify the most likely values for the problem and then we compute the query scores, originated parts of values for M and E, and which are the most likely values for MMD. Another approach we experimented with consists in directly optimizing the vector representations associated with the variables MMD. This can be done by a gradient based optimization. We first initialize the embeddings of M and the randomly. And then we optimize the embedding of them and the to maximize the score of the query. And finally replaced the embedding of them with the embedding of all the candidate entities and rank them using the corresponding query score',\\
\hl{'Results':} 'Yeah. So we experimented on a variety of complex grid structures. And despite being only trained on simple creation, we can see that our model systematically generalize is too complex queries better than models trained on complex queries. In the first place, He had results on 3 - 15 K 237 for different complex great types. Here we can see the results on three basic 15 K. Any other results on 995 and here we have the average results between all types of complex queries that we considered. And we can see that our model produces significantly more accurate results on all the traii datasets. This improves of the existing models, both in terms of the data efficiency, because we only need to train our model on a much smaller dataset of simple queries and also in terms of out of out of distribution generalization, because we get better generalization accuracy on complex queries without having to train on them In the 1st place. Another really nice feature of this model is that it can provide explanations for its predictions. Other models in the space only returns a list of answers to the query. For instance, in this case, the answer is being produced at a pixel ban, I'm 15 block setting and others. Our model can also be used to provide the intermediate results associated with each query in the form of the variable assignments used to produce the answer. For example, here we can see that a pixel ban and I'm tellin were considered as answers because according to the model, they interact with oxytocin while dual oxygen was considered as as an answer because according to the model, it interacts with pregabalin and this allows us to check whether the results are being produced for the right reasons. In this case box setting and pregabalin are two possible treatments for anxiety. So the model is producing the correct answers for the correct reasons. This is not always the case. For example, consider the following query from feedback is 15 K 237. What international organizations contain the country of nationality of thomas Aquinas here, the model was able to return the correct set of answers NATO asI, D U and Wt O. However, this return for the wrong reasons. Um thomas Aquinas was mistakenly assumed to be from the US from the UK are from Germany. While the correct nationality of thomas Aquinas is italian. Our model enables us to detect such iris and possibly cracked them by refining the underlying newly prediction model.'}
\end{tcolorbox}
\caption{Presentation Example - 2}
\label{oral-2}
\end{figure*}

\begin{figure*}
\begin{tcolorbox}[enhanced,attach boxed title to top center={yshift=-3mm,yshifttext=-1mm},boxrule=0.9pt,colback=gray!00,colframe=black!50,colbacktitle=ProcessBlue!70,title=Presentation Example - 3,boxed title style={size=small,colframe=ProcessBlue!70},sharp corners,left=5mm,right=5mm,top=2mm,bottom=2mm,boxsep=0pt]
{\small
\hl{'Ending':} 'So, to summarize in this paper, we propose a new approach France and complex creates on large scale and incomplete knowledge graphs. In our approach, we first train a neural prediction model on the task of answering simple atomic creation. And then we cast the problem of answering complex grace as an optimization problem where we need to find the mapping from variables, quantities that maximizes the score of the complex square. And we show that our approach generalizes extremely well to complex queries despite not having been trained on them in the first place, all the source code, the pre trained models and datasets online at this link. And if you want to collaborate on this topic or have any questions or comments, please feel free to reach out to us and thank you for listening. thank you very much So you may be. Now we will move to the next paper about complex query answering',\\
\hl{'Q\&A':} 'So we had one question from Cartier uh she mentioned that the work is supplied to relatively small data said she wants to know essentially what happens if you try to scale it up. And I follow with the second question quickly. I was wondering essentially because you have these different types of you know, complex queries and then you have two types of optimization. One is the continuous relaxation, another is the greedy discrete approach. Which one is better in which case is there any kind of ideal correlation between types of I don't know, complexity of the queries, how they are designed and so on or Yes, thank you. Thanks to the great questions. So about the first one about getting the method up to much larger quantities of data like the millions um of data points they use in other methods. I think it's super interesting and we can I think we might observe significant improvement for that from that. For example. One thing that we're doing at the moment is we are fixing how so we select one continuous relaxation of the logical end and are as an input parameter, we select the two norm and that economic beforehand as an input parameter and then we just execute the method but I think by scaling this matter to larger conditional data and to complex creates during training we can think of for example to um to train how we represent the logical and and or within the architecture and we can we would be able to use for example, paramedic economics and economies, which we are not touching at the moment because we don't use complexities, complexities during training. And also um the naturally predictors that we use for answering economic queries are not really trained. Um so um are not really trained on complex crises in in the first place that trained on atomic quiz. And I think that so the scores are not really trying to do interact together in some sense. So I think we might be able to observe a significant improvement from that as well. About the other question. So what we observed is that the discrete search seems to work consistently better than the continuous search um across all datasets and the types of complex queries. And we think that that might happen because the the continuous search for the entity representations might find some entity representations that do not do not correspond to any real entities in some sense. Or maybe the model might hallucinate some, I don't know, a dog with seven legs. Uh The other entities that are not really close in terms of representations to real entities in the knowledge base. So I think that's my that that might be a reason why um this could search works consistently better on uh continue such. Um Thanks. Okay, thank you very much. I think that answers the question. So I'll hand over tell me actually have another question about the linking paper. So you have a two step approach for your first predict links and then you run queries on top of that. Can you comment on? No, no, basically the links are predicted. Um so basically the links are predicted as kind of a part of the optimization problem. We translate the complex series into each complex query is translated into an optimization problem and the neural predictor. Um that makes whether there is a link between two entities is kind of a component of this optimization problem because correcting like all drinks beforehand, doesn't scale up because you can have billions of possible links uh in the name, like materializing all the links beforehand doesn't really uh scale also like you need to decide uh what whether to materialize a link or not. So basically the link prediction process is part of the complex variants diagnosis. Uh this map'.}
\end{tcolorbox}
\caption{Presentation Example - 3}
\label{oral-3}
\end{figure*}
\end{document}